\documentclass[lettersize,journal]{IEEEtran}
\usepackage{amsmath,amsfonts}
\usepackage{algorithmic}
\usepackage{algorithm}
\usepackage{array}
\usepackage[caption=false,font=normalsize,labelfont=sf,textfont=sf]{subfig}
\usepackage{textcomp}
\usepackage{stfloats}
\usepackage{url}
\usepackage{verbatim}
\usepackage{graphicx}
\usepackage{cite}

\usepackage{amsmath,amsfonts,bm}

\def\eqref#1{equation~\ref{#1}}
\def\1{\bm{1}}

\DeclareMathAlphabet{\mathsfit}{\encodingdefault}{\sfdefault}{m}{sl}
\SetMathAlphabet{\mathsfit}{bold}{\encodingdefault}{\sfdefault}{bx}{n}

\usepackage{hyperref}
\usepackage{url}
\usepackage{enumitem}
\usepackage{color}
\usepackage{xspace}
\usepackage{amsmath}
\usepackage{graphicx}
\usepackage{multirow}
\usepackage{enumitem}
\usepackage{algorithm}
\usepackage{algorithmic}
\usepackage{amsthm}
\usepackage{amssymb}
\usepackage{amsfonts}
\usepackage{diagbox}
\usepackage{wrapfig}
\usepackage{xcolor}

\newcommand{\acc}[2]{#1\tiny$\pm$#2}

\newcommand{\FStransfer}{\textsc{FewTrans}\xspace}
\newcommand{\keypoint}[1]{\vspace{0.1cm}\noindent\textbf{#1}}

\begin{document}

\title{\textbf{Benchmarking Few-shot Transferability of Pre-trained Models 
with Improved Evaluation Protocols}}

\author{Xu~Luo,
        % Jing~Xu,
        Ji Zhang$^{\dagger}$,
        Lianli~Gao,
        % Zenglin~Xu,
        % Difan~Zou,
        Heng Tao~Shen
        and~Jingkuan~Song% <-this % stops a space  $^{\dagger}$
\IEEEcompsocitemizethanks{
\IEEEcompsocthanksitem Xu Luo and Lianli Gao are with University of Electronic Science and Technology of China, Chengdu, China.
E-mail: frank.luox@outlook.com, lianli.gao@uestc.edu.cn
\IEEEcompsocthanksitem Ji Zhang is with Southwest Jiaotong University, Chengdu, China. 
E-mail: jizhang.jim@gmail.com
\IEEEcompsocthanksitem Heng Tao Shen and Jingkuan Song are with Tongji University, Shanghai, China.
E-mail: shenhengtao@hotmail.com, jingkuan.song@gmail.com
\IEEEcompsocthanksitem $^{\dagger}$ Ji~Zhang is the corresponding author.
}
% <-this % stops an unwanted space
}

\maketitle

\begin{abstract}
Few-shot transfer has been revolutionized by stronger pre-trained models and improved adaptation algorithms. However, there lacks a unified, rigorous evaluation protocol that is both challenging and realistic for real-world usage. In this work, we establish \FStransfer, a comprehensive benchmark containing 10 diverse datasets, and propose the Hyperparameter Ensemble (HPE) protocol to overcome the ``validation set illusion'' in data-scarce regimes. Our empirical findings demonstrate that the choice of pre-trained model is the dominant factor for performance, while many sophisticated transfer methods offer negligible practical advantages over a simple full-parameter fine-tuning baseline. \textcolor{black}{To explain this surprising effectiveness, we provide an in-depth mechanistic analysis showing that full fine-tuning succeeds via distributed micro-adjustments and more flexible reshaping of high-level semantic representations without suffering from overfitting. Additionally, we quantify the performance collapse of multimodal models in specialized domains as a result of linguistic rarity using adjusted Zipf frequency scores.} By releasing \FStransfer, we aim to provide a rigorous ``ruler'' to streamline reproducible advances in few-shot transfer learning research. \textcolor{black}{We make the \FStransfer benchmark publicly available at \url{https://github.com/Frankluox/FewTrans}.}
\end{abstract}

\section{Introduction}

Recent progress in computer vision \cite{BiT,CLIP,islam2021broad,dehghani2023scaling} suggests that good performance on a variety of vision tasks can be achieved at low cost by transferring a pretrained, large-scale model with only a few labeled samples, facilitating downstream scenarios where labeled data can be expensive or difficult to obtain. This \emph{few-shot transferability} of pre-trained models can be further improved by adopting recently proposed transfer algorithms that are claimed to be better than vanilla finetuning in terms of accuracy or efficiency, such as partial finetuning \cite{BitFit}, low-rank adaptation \cite{LoRA}, adapter tuning \cite{Adapter,wu2024fine,zhang2025reliable,TSA}, meta-learning \cite{tian2022adversarial,zhang2022progressive,zhang2022few}, prompt tuning \cite{CoOP,Maple,zhang2025closer,zhang2023channel} and so on.

However, the evaluation criteria of few-shot transfer have not been unified and diverge across separate threads of research, which hinders newly proposed pretrained models or transfer algorithms from being accurately evaluated and compared with previous ones. Through systematic experiments, we identify two critical deficiencies in previous evaluation setups caused by the specific nature of few-shot problems. First, we observe a ``sampling lottery'' effect, where extreme performance variation caused by random task sampling makes reports based on few tasks highly unreliable. Second, we highlight the ``validation set illusion'': traditional hyperparameter selection often relies on large, target-domain validation sets that are impractical in true few-shot scenarios. Our analysis reveals that optimal hyperparameters vary significantly across tasks and datasets, making it exceptionally difficult to design a fair model selection criterion. To address this, we propose the Hyperparameter Ensemble (HPE) protocol \cite{wenzel2020hyperparameter}. Unlike traditional cross-validation, which fails under extreme data scarcity, HPE leverages a range of hyperparameters for prediction, providing a robust and label-free evaluation that captures the true potential of an algorithm.

Integrating all our solutions, we construct \FStransfer, a benchmark containing 10 diverse downstream datasets with the ability of sampling class-imbalanced tasks with varying numbers of classes and shots. \textcolor{black}{We demonstrate that our HPE protocol naturally penalizes volatile methods through a positive correlation ($r=0.38$) between sensitivity and ensemble penalty, ensuring a fair evaluation that accurately reflects real-world deployment risks. Beyond identifying evaluation flaws, this work provides a mechanistic understanding of few-shot adaptation. We observe that while a larger pre-training dataset contributes significantly to the downstream performance, the gaps between sophisticated transfer algorithms are often negligible. Notably, simple all-parameter fine-tuning (Full-FT) performs surprisingly well and avoids the expected overfitting. We provide an in-depth analysis showing that Full-FT succeeds through distributed ``micro-adjustments'' that preserve pre-trained knowledge while effectively reshaping deep semantic features. Furthermore, we quantify the performance collapse of multimodal models on specialized domains as a consequence of linguistic rarity using adjusted Zipf frequency scores. Finally, paired statistical tests across 6000 tasks reveal that the practical advantages of sophisticated transfer algorithms over simple fine-tuning are often negligible, providing a sobering look at the true state of progress in the field.}

\textbf{The main contributions of this work are threefold:}
\begin{itemize}
\item We introduce FEWTRANS, a challenging new benchmark for few-shot transfer learning that establishes a more rigorous and realistic evaluation paradigm, eliminating the sampling lottery through large-scale task reporting.
\item We propose the hyperparameter ensemble protocol as a robust evaluation method to overcome the performance instability and impracticality of traditional tuning, ensuring fairness without requiring additional labels.
\item \textcolor{black}{We provide mechanistic insights into few-shot transfer, explaining the efficacy of full fine-tuning via parameter update scales and identifying linguistic domain shift as the primary bottleneck for multimodal model adaptation.}
\end{itemize}

\color{black}
\section{Related Work}

\keypoint{Pre-training paradigms.} The few-shot transferability of models is fundamentally driven by scaling training data, model architectures, and pre-training objectives. Early studies verified that supervised ImageNet models provide significantly better downstream performance than training from scratch \cite{kornblith2019better}, while subsequent research demonstrated that self-supervised models often serve as superior source models \cite{islam2021broad,luo2023closer}. Recent scaling laws suggest that increasing pre-training datasets to hundreds of millions and parameters to billions leads to consistent improvements in few-shot transfer \cite{BiT,zhai2022scaling,wang2025dualbranchpromptingmultimodalmachine}. Additionally, multimodal models like CLIP \cite{CLIP} leverage natural language supervision to achieve zero-shot and few-shot capabilities across diverse visual tasks.

\keypoint{Transfer algorithms.} To effectively adapt large-scale models, various algorithms have been proposed beyond vanilla fine-tuning. These include parameter-efficient fine-tuning (PEFT) methods such as BitFit \cite{BitFit}, LoRA \cite{LoRA}, and various adapter-based tuning techniques \cite{Adapter,wu2024fine,TSA}. For multimodal models, prompt tuning methods like CoOp \cite{CoOP} and MaPLe \cite{Maple} focus on optimizing soft prompts rather than model weights. Broader research has also extended few-shot adaptation to specialized domains, such as infrared imagery and object detection \cite{zhang2025benchmark}, utilizing techniques like dual-domain feature extraction \cite{zhang2021deep}, antagonistic learning \cite{zhang2023differential}, or part-aware correlation networks \cite{zhang2024part}. While these methods aim for high efficiency, our benchmark reveals that their advantages over simple full-parameter fine-tuning are often negligible under rigorous evaluation.

\keypoint{Evaluation protocols.} Standard many-shot benchmarks like VTAB \cite{VTAB} are well-established, but few-shot transfer evaluation remains fragmented. Unlike previous works that rely on self-selected datasets \cite{kornblith2019better,CLIP} or single-seed reports without confidence intervals, \FStransfer mandates many-task sampling to eliminate the ``sampling lottery'' \cite{sensitivity}. Furthermore, most existing protocols suffer from the ``validation set illusion,'' where hyperparameters are tuned on large target-domain validation sets that are unavailable in real-world few-shot scenarios \cite{luo2023closer}. While Meta-Dataset \cite{meta-dataset} established early standards for few-shot classification, it remains unsuitable for foundation model transfer due to the absence of class names, unnatural image preprocessing, and excessive sample counts. \FStransfer explicitly contrasts with these practices by introducing a novel Hyperparameter Ensemble (HPE) protocol, ensuring a realistic and label-free evaluation of model transferability.

\color{black}

\begin{figure}[t]
% \vskip 0.2in
\centering
\centerline{\includegraphics[width=0.8\linewidth]{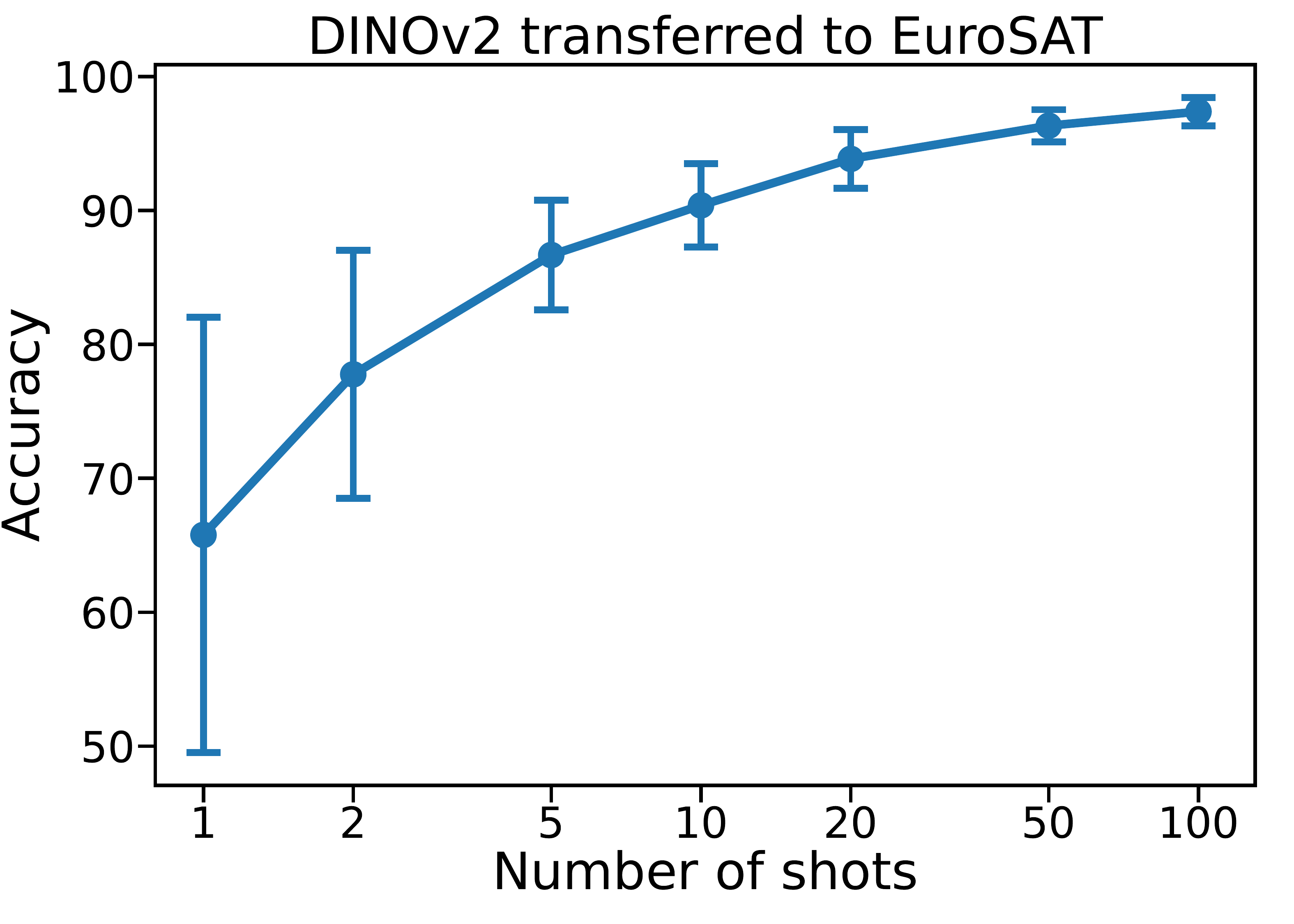}}
\caption{Average accuracy and 95\% confidence intervals of single-task few-shot transfer evaluation  of a pretrained DINOv2-s model on EuroSAT \cite{eurosat}.}
\label{shot_variance}
\vskip -0.in
\end{figure}

\section{The Problem of Few-shot Transfer Learning}
In transfer learning, we have a pretrained model $f_{\theta}: \mathbb{R}^d\rightarrow\mathbb{R}^m$ mapping inputs $x\in\mathbb{R}^d$ to features $z\in\mathbb{R}^m$. The goal of transfer learning is to transfer the pretrained model $f_{\theta}$ to a specified downstream task. Any downstream task $\tau$ can be described as a combination of a training set $D^{tr}=\{(x_i,y_i)\}_{i=1}^N$ and a test set $D^{te}=\{(x^*_j,y^*_j)\}_{i=1}^M$, where $y_i, y^*_j\in\{1,...,n_{cls}\}$ are class labels. The task $\tau$ is called a $K$-shot task if there are exactly $K$ samples per class in $D^{tr}$. During transfer, the pretrained model $f$ will be adapted to task $\tau$ using the training set $D^{tr}$ through a transfer algorithm such as finetuning, producing a new classifier mapping images to labels of the new task. To evaluate the effectiveness of transfer, the produced classifier will be evaluated on the test set $D^{te}$. 

In a typical transfer learning evaluation setup \cite{VTAB}, the downstream task involves an entire downstream dataset, so the number of samples per class can be quite large, deviating from some practical transfer scenarios where downstream data is difficult to obtain. Under few-shot transfer scenario, the number of samples per class can be quite small, usually less than 20 or 10.

\section{Inappropriate Evaluation of Previous Methods}
\label{sec:Inappropriate}
In this section, we point out several flaws of previous few-shot transfer evaluation protocols. For all experiments done in this section, we use fine-tuning as the transfer algorithm. Following \cite{luo2023closer}, we separately set the learning rates for the backbone of the pretrained model and the linear head for improved performance. By default, we use Adam \cite{adam} as the optimizer.

% \subsection{Preliminaries}

% \begin{wrapfigure}[17]{l}{0.45\textwidth}
%      \centering 
%      % \vspace{-12pt}
% \includegraphics[width=0.45\textwidth]{Images/shot_analysis.pdf}
% % \vspace{-20pt}
%      \caption{Average accuracy and 95\% confidence intervals of single-task few-shot transfer evaluation  of a pretrained DINOv2-small model on EuroSAT \cite{eurosat}.} 
%      \label{shot_variance}
% \end{wrapfigure}

% \begin{figure}[t]
% % \vskip 0.2in
% \centering
% \centerline{\includegraphics[width=0.6\linewidth]{Images/shot_analysis.pdf}}
% \caption{95\% confidence intervals of few-shot transfer performance of pretrained DINOv2-small model on EuroSAT \cite{eurosat}. The variation can be very large when the number of shots is small.}
% \label{shot_variance}
% \end{figure}

\subsection{Large Performance Fluctuation Caused by Sampling}
Different from the typical transfer learning setup where an entire dataset is used as the downstream task, in few-shot transfer, a randomly sampled small part of the dataset is used as the downstream task. Previous works that evaluate pretrained models on few-shot transfer tasks \cite{BiT,CLIP,CoOP} sample a single or a few (usually 3) tasks and only report the average performance on the sampled tasks without error bars. This can be problematic because the performance can be largely influenced by the choice of sampled data especially under few-shot settings. To illustrate this, we give the single-task transfer performance of DINOv2-small \cite{Dinov2} on the EuroSAT dataset \cite{eurosat} along with the 95\% confidence intervals in Figure \ref{shot_variance}. As seen, when the number of shots is small, the  spread of the error bar can be very large. For 1-shot task, the performance can vary from less than 50\% to more than 80\% within the confidence interval. This is caused by the randomness of the training set, where a change to a single sample can lead to large fluctuations in performance \cite{sensitivity}. Thus the comparison in previous works using only a few tasks is inappropriate because the change of seed can determine the rank of pretrained models/transfer methods completely. To make the comparison meaningful, we should at least sample more tasks to make the confidence interval small enough.

\subsection{Unrealistic Model Selection}
In the typical transfer learning setting, the downstream dataset is so large that we can partition it into a training set for adptation, and a validation set for selecting hyperparameters like learning rates and number of epochs for adaptation. When it comes to the evaluation of few-shot transferability of pretrained models, previous works either tune hyperparameters on a large validation set (possibly from different classes) from the same dataset~\cite{CLIP,luo2023closer} or set hyperparameters to empirically-derived default values dependent on downstream datasets~\cite{BiT,CoOP}. While it seems valid to tune hyperparameters on a separated validation set as what is done in the traditional many-shot transfer learning literature, we point out that doing so is inappropriate in the few-shot setting because it deviates from real-world scenarios where additional labeled data from the same dataset for validation is hard to obtain.

Thus to make the evaluation protocol realistic while being fair for comparison, we have two choices: (1) determine hyperparameters of transfer algorithms in advance on a \emph{held-out dataset} that is both different from the pretraining dataset and target downstream dataset; (2) determine hyperparameters based on the few training samples of the target downstream dataset on the fly. We will next evaluate the validity of these two choices.

\begin{table*}[t]
% \footnotesize
% \scriptsize
\centering
\setlength\tabcolsep{10pt}
\caption{Optimal hyperparameters vary from task to task. The pretrained model is DINOv2-small, and all tasks are 1-shot sampled from EuroSAT.}
% \linespread{1.5}
\begin{tabular}{c|cccccccccc}
\hline
Task ID &0& 1& 2& 3 &4 &5 &6& 7 &8& 9\\\hline
Epoch & 15 & 15 & 15 & 40 & 15 & 40 & 30 & 20 & 20 & 30\\
Backbone lr & 5e-05 & 5e-06 & 5e-06 & 1e-05 & 2e-06 & 5e-06 & 1e-05 & 2e-05 & 2e-05 & 1e-05\\
Head lr & 0.05 & 0.01 & 0.2 & 0.02 & 0.01 & 0.05 & 0.01 & 0.05 & 0.05 & 0.2\\
\hline
\end{tabular}
\label{vary_from_task_to_task}
\end{table*}

\subsubsection{\textbf{Optimal hyperparameters change from task to task}} If we determine hyperparameters on a separate dataset, then the hyperparameters will be the same for all tasks. Is this appropriate? In Table \ref{vary_from_task_to_task}, we show the optimal hyperparamters of ten tasks sampled from the same dataset.
We can observe that, even when sampled from the same dataset with the same set of classes, tasks with different training samples can have different optimal hyperparamters. The optimal number of epochs varies from $15$ to $40$; the optimal learning rate for pretrained backbone varies from $2e-06$ to $5e-05$; the optimal learning rate for linear classifier varies from $0.01$ to $0.2$. 

\subsubsection{\textbf{Few-shot transfer performance is sensitive to the choice of hyperparameters}} 
\begin{figure*}[t]
% \vskip -0.2in
\centering
\centerline{\includegraphics[width=0.8\linewidth]{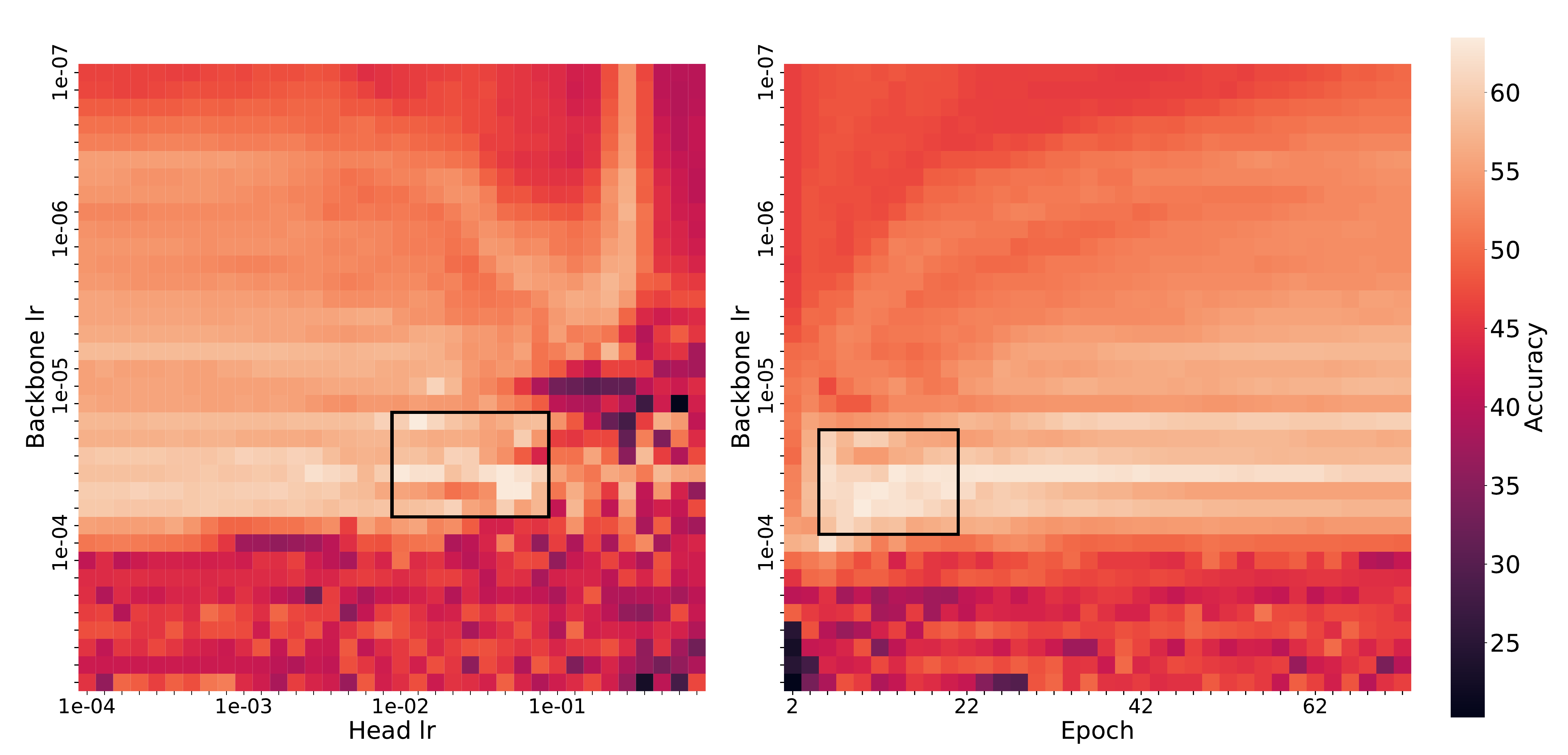}}
\caption{The heatmaps showing how the few-shot transfer performance of a single 1-shot task sampled from EuroSAT changes with hyperparameters. We fix the number of epochs to $50$ in the left plot, and fix the head lr to $0.01$ in the second plot. The black rectangles highlight the optimal hyperparameter areas.}
\label{one_task_visualization}
\vskip -0in
\end{figure*}

Only showing that the optimal hyperparameters change from task to task is not enough to conclude that the few-shot transfer performance will change from task to task if we use the same hyperparameters for all tasks. We still need to show that few-shot transfer performance is sensitive to hyperparameters. We show how sensitive few-shot transfer performance is to the choice of hyperparameters in Figure \ref{one_task_visualization}. We plot the heatmaps of few-shot transfer performance of a single 1-shot task when varying two of the hyperparameters. As we can see, the variation of accuracy can be very large in the considered ranges, from around $20$\% to more than $60$\%. In particular, the performance can drop very quickly when we move out of the optimal area (highlighted in the black rectangle). For example, in the left plot, if we go down or right from the black rectangle, that is, increasing the learning rate of the backbone or the linear head, we will go into a chaotic area, where the accuracy oscillates up and down irregularly and often drops to half or even less. This phenomenon seems not that evident for the number of epochs in the right plot where the accuracy seems to be smoother, but we can still see a $10$\% performance fluctuation around the optimal area.

\subsubsection{\textbf{Optimal hyperparameters change from dataset to dataset}} Even if we can tolerate the performance variation per task, we show that the ``average optimal hyperparameters''---the hyperparameters that give the highest average performance over several tasks sampled from a dataset---can still vary from dataset to dataset in Table \ref{dataset_to_dataset}. For example, the optimal number of epochs when transferred to Plant Disease \cite{PlantD} is $50$, while the optimal number of epochs when transferred to UCF101 \cite{ucf101} is $10$. Among the six downstream datasets, the backbone learning rate ranges from $1e-06$ to $2e-05$, and the head learning rate ranges from $5e-04$ to $1e-02$.

Combining the analysis above, we can conclude that single hyperparameters will lead to unstable few-shot transfer performance from task to task and from dataset to dataset. So this hyperparameter selection criteria will cause large uncertainty of few-shot transfer performance and thus cannot reflect the true performance of different methods. Thus a proper hyperparameter selection criteria should rely only on the training set of the downstream dataset at hand, which we will explore next.

\subsubsection{\textbf{Cross-validation fails to provide reliable estimation of hyperparameters}} 
\begin{figure*}[t]
% \vskip 0.2in
\centering
\centerline{\includegraphics[width=0.65\linewidth]{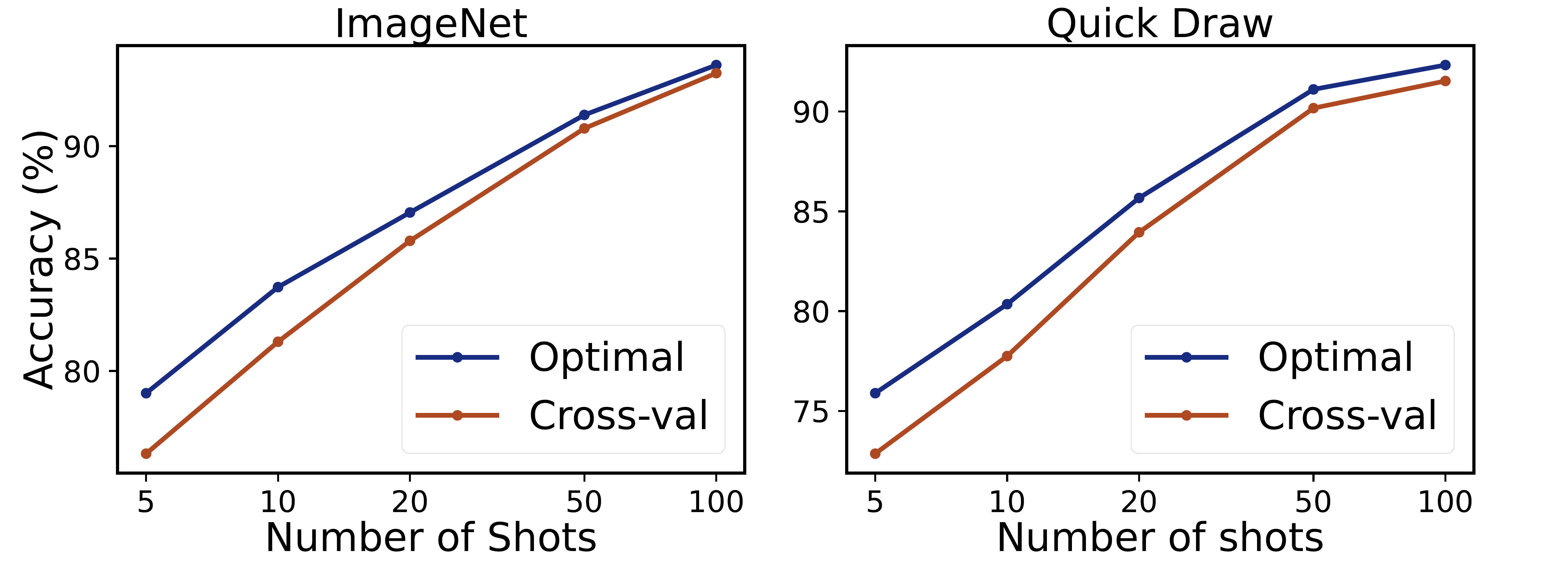}}
\caption{Cross-validation cannot find good hyperparameters when the number of shots is small, regardless of the domain shift between pretraining and downstream dataset. We use a subset class of ImageNet as the training set, and use the remaining part as the downstream dataset for the left plot.}
\label{optimal_vs_crossval}
\vskip -0in
\end{figure*}

A representative way of estimating the hyperparameters using the training set of downstream tasks is \emph{cross-validation} which has a long history of use in machine learning \cite{kohavi1995study,cross_validation}.
The main idea behind $l$-fold cross-validation is to split data $l$ times, each time into a training part and a validation part. For each time of split, the training part is used to adapt the model and the validation part is used to evaluate the adaptation. The hyperparameters are chosen such that the average error over all splits is small. 
While cross-validation can work well when there is abundant data, we find that it will meet difficulties when data for adaptation is scarce, because (1) the number of samples per class is too small to split. For example, when the number of samples is below $5$, it is only possible to use the leave-one-out strategy, that is, the validation part only has $1$ sample for each spilt, leading to unreliable performance estimation. For extreme $1$-shot case, we cannot apply cross-validation because there is no data to split; (2) $l$-fold cross-validation changes the number of shots from $K$ to $K(l-1)/l$. As we have shown previously, optimal hyperparameters for few-shot transfer can change when the task has changed, thus the hyperparameters found by cross-validation can be biased. We verify our considerations in Figure \ref{optimal_vs_crossval}, where we show that there is a gap between the accuracy obtained by 5-fold   cross-validation and the accuracy obtained by using the ``average optimal hyperparameters'' of the dataset for both in-domain and out-of-domain transfer, especially when the number of shots is small.

In conclusion, figuring out the optimal hyperparameters for few-shot transfer is very important and is, if not impossible, very difficult under real-world settings. Because of this difficulty, a good few-shot transfer method should not only have high performance at its optimal hyperparameters, but should also have resistance to the change of hyperparameters, that is, the test loss landscape around the optimal hyperparameters should be flat such as we can tolerate an inevitable deviation of hyperparameter estimations. Thus a good evaluation protocol should evaluate both the performance that a pretrained model can reach, as well as its sensitivity to the choice of hyperparameters.

\begin{table}[t]
% \footnotesize
% \scriptsize
\centering
\setlength\tabcolsep{2pt}
\caption{Average optimal hyperparameters of few-shot transfer vary from dataset to dataset. The pretrained model is DINOv2-small, and all tasks are 1-shot.}
\begin{tabular}{c|cccccc}
\hline
 &CIFAR-100& UCF & Plant Disease & Aircraft & DTD &EuroSAT\\\hline
Epoch & 30 & 10 & 50 & 30 & 20 & 30\\
Backbone lr & 1e-05 & 1e-05 & 2e-05 & 1e-06 & 1e-05 & 1e-05\\
Head lr & 0.0005 & 0.01 & 0.005 & 0.005 & 0.001 & 0.01\\
\hline
\end{tabular}
\label{dataset_to_dataset}
\end{table}

\subsection{Other Considerations}
Apart from the aforementioned two major defects of previous evaluation of few-shot transferability of pretrained models, we also notice several other points that can be improved, with some of them inspired by the few-shot learning literature \cite{meta-dataset}.

\subsubsection{\textbf{No variation of the number of classes}}
Following transfer learning literature, papers that evaluates the few-shot transferability of pretrained models often use all classes of the target downstream dataset \cite{BiT,CLIP} to form a task, thus the capability of pretrained models transferring to less number of classes which forms more specific fine-grained category structures is not considered.

\subsubsection{\textbf{No class imbalance}} For simplicity, almost all previous few-shot transfer evaluations use class-balanced settings, where the number of shots in each class of the training set is exactly the same for all classes. However, we cannot guarantee that this will still hold in real-world few-shot transfer scenarios and thus models and algorithms should be evaluated on class-imbalanced scenarios.

\subsubsection{\textbf{Datasets lack diversity, are too easy, and may have errors}} Take the widely-used few-shot transfer benchmark for multimodal pretrained models \cite{CLIP,CoOP} that contains $11$ datasets as an example. Images from most datasets in this benchmark are taken from modern cities, thus being similar to parts of ImageNet and the tasks are not difficult to solve even when there are only a few samples per class. This can be seen from recent papers \cite{Maple}
where the average few-shot accuracy of 5 datasets reaches more than 90\%, in the condition that some of the datasets have more than 100 classes, which should have been difficult to classify correctly with few samples per class. In addition, the benchmark has StanfordCars \cite{stanfordcars} as one of its datasets, which has proven to have tons of mislabeled images and outliers \cite{stanforderror}.

\section{Introducing the FewTrans Benchmark}
In this section, we introduce several evaluation standards to solve the aforementioned issues of few-shot transfer, which constitutes the key components of our proposed \FStransfer benchmark. 

\color{black}
\subsection{Hyperparameter Ensemble for Robust Few-shot Evaluation}
\label{sec:hyperparameter_ensemble}

Selecting optimal hyperparameters (e.g., learning rates, training epochs) is notoriously difficult in few-shot scenarios due to the ``validation set illusion''---the inherent lack of a sufficiently large validation set to reliably estimate performance. Traditional cross-validation often yields biased estimates or fails entirely in extreme 1-shot cases. To address this, we propose the Hyperparameter Ensemble (HPE) protocol, which circumvents the search for a single ``best'' configuration by aggregating predictions from a range of candidates.

\keypoint{Formal definition and fusion strategy.} Let $\mathcal{T}$ be a few-shot task with a support set $\mathcal{S}$ and a query set $\mathcal{Q}$. We define a predefined grid of $N$ hyperparameter configurations as $\mathcal{H} = \{h^{(1)}, h^{(2)}, \dots, h^{(N)}\}$. For each configuration $h^{(i)} \in \mathcal{H}$, the transfer algorithm adapts the pre-trained model using the support set $\mathcal{S}$, resulting in an adapted classifier $g_{\phi_i}: \mathcal{X} \to \mathbb{R}^C$. For any test sample $x \in \mathcal{Q}$, the final ensemble classification score $S(x)$ is defined as the average of the logits produced by all individual configurations:
\begin{equation}
    S(x) = \frac{1}{N} \sum_{i=1}^N g_{\phi_i}(x).
\end{equation}
The final predicted label is determined by $\hat{y} = \arg\max_c S_c(x)$. This strategy allows the evaluation to capture high-performance signals within the hyperparameter space while recovering ``Oracle'' performance without an explicit validation set.

\begin{table*}[t]

% \footnotesize
% \scriptsize
\centering
\setlength\tabcolsep{4pt}
\caption{Average 1-shot transfer  performance of pretrained DINOv2-small over 50 tasks: hyperparameter ensemble vs. individual hyperparameter configurations. See appendix for details.}
\begin{tabular}{c|ccccccccc|ccc}
\hline
configuration & (1,1)& (1,2)& (1,3)& (2,1)& (2,2)& (2,3)& (3,1)& (3,2)& (3,3) & Avg & lr ensemble& lr+epoch ensemble\\\hline
EuroSAT & 67.52 & 67.23 & 68.00 & 70.75 & 71.01 & 61.87 & 45.44 & 45.55 & 42.48 & 59.98 & 70.21&70.72\\
Aircraft & 61.92 & 61.55 & 61.44 & 61.49 & 61.79 & 61.23 & 61.44 & 61.31 & 60.45 & 61.40 & 63.39 &63.28\\
\hline
\end{tabular}
\label{hyperparameter_ensemble}
\end{table*}

\keypoint{Robustness to Individual Configurations.} A primary advantage of HPE is its robustness to individual bad configurations. As demonstrated in Table \ref{hyperparameter_ensemble}, even when some configurations perform poorly (e.g., yielding accuracies below 50\% on EuroSAT), the ensemble maintains a high performance of 70.21\%, remarkably close to the optimal individual configuration (71.01\%). As long as the search range covers the optimal region, the evaluation is stabilized against the ``sampling lottery.''

\begin{figure*}[t]
% \vskip 0.2in
\centering
\centerline{\includegraphics[width=0.8\linewidth]{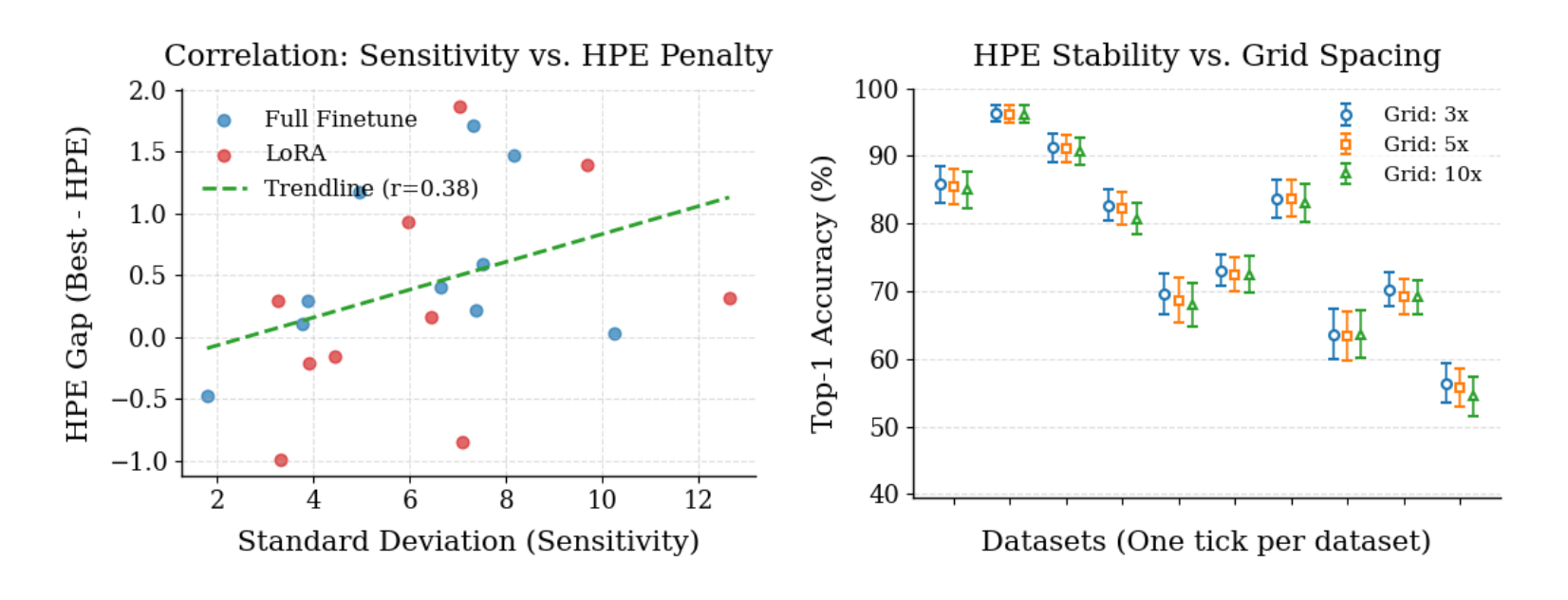}}
\caption{Robustness and fairness analysis of the Hyperparameter Ensemble (HPE) protocol. Left: Positive Pearson correlation ($r=0.38$) between hyperparameter sensitivity and the HPE penalty, confirming that the protocol naturally penalizes volatile methods. Right: Stability of HPE Top-1 accuracy across various grid spacings (3x, 5x, 10x), demonstrating that the protocol effectively buffers against the specific choice of hyperparameter boundaries across all 10 datasets.}
\label{fig: tmm_rebuttal_2}
\vskip -0in
\end{figure*}

\keypoint{Sensitivity, Fairness, and Penalty.} Beyond stabilizing performance, HPE serves as a metric for the sensitivity of an algorithm. When the test loss landscape around the optimal region is flat, the ensemble gain is more consistent. Conversely, the protocol naturally incorporates a sensitivity penalty: volatile methods are ``pulled down'' by their poor-performing configurations in the ensemble. This accurately reflects the risk of such methods in real-world scenarios where validation data is unavailable. Our analysis reveals a positive Pearson correlation ($r=0.38$) between an algorithm's hyperparameter sensitivity and the resulting HPE penalty (the gap between Oracle-best and HPE scores), as shown in Figure \ref{fig: tmm_rebuttal_2}. This confirms that HPE effectively penalizes volatile methods rather than masking their unreliability. To ensure fairness, all baselines utilize the same $3\times3$ grid protocol ($N=9$), measuring intrinsic robustness rather than capacity under cherry-picked conditions.

\keypoint{Stability across Grid Spacings.} We further evaluate HPE's robustness to the choice of hyperparameter boundaries. As illustrated in Figure \ref{fig: tmm_rebuttal_2}, the Top-1 accuracy remains remarkably stable across all 10 datasets regardless of grid spacing (3x, 5x, or 10x the center learning rate). This stability proves that the HPE protocol effectively ``buffers'' against the specific choice of grid boundaries, providing a consistent and reliable evaluation ruler for the community.

\noindent\textcolor{black}{\textbf{Remark.} Empirically, we demonstrate that HPE consistently approximates the individual Oracle best configuration performance. However, it is important to explicitly acknowledge that the rigorous theoretical proof of logit averaging as a perfect universal approximation to the Oracle selector in non-convex few-shot landscapes remains an open question for the community. By providing this empirical ``ruler'', we hope future work can stimulate further theoretical investigation into the mechanisms of HPE under extreme data scarcity.}

\color{black}
\subsection{Other components of \FStransfer}

\color{black}
\subsubsection{Datasets and Selection Criteria}
The construction of the \FStransfer suite followed a structured filtration process from an initial pool of over 40 representative candidate datasets sourced from VTAB \cite{VTAB}, Meta-Dataset \cite{meta-dataset}, and multimodal transfer literature such as MaPLe \cite{Maple}. To ensure the benchmark's uniqueness, scientific rigor, and challenging nature, we applied four critical filters:

\begin{itemize}[leftmargin=*]
    \item \textbf{Multimodal compatibility:} Every dataset must have natural language class names to support vision-language models like CLIP. This necessitated the exclusion of various datasets from VTAB and Meta-Dataset that utilize abstract or numeric identifiers.
    \item \textbf{Domain redundancy reduction:} We maximized semantic diversity by selecting single representatives from overlapping domains (e.g., choosing CIFAR-100 \cite{CIFAR} while excluding CIFAR-10).
    \item \textbf{Exclusion of saturated tasks:} We excluded ``easy'' tasks where modern pre-trained models consistently achieve near-saturated performance, such as Oxford Pets \cite{oxfordcat}, Food101 \cite{bossard2014food}, and Caltech101 \cite{caltech}, as they fail to provide a meaningful algorithmic challenge.
    \item \textbf{Data integrity:} Datasets with known high label noise or significant outliers, most notably Stanford Cars \cite{stanfordcars}, were removed to ensure result reliability.
\end{itemize}

Following this filtration, we selected ten datasets that cover a diverse spectrum of domains: ImageNet-Sketch \cite{ImageNet_sketch}, DTD \cite{DTD}, CIFAR-100 \cite{CIFAR}, VGG Flowers \cite{VGG}, UCF-101 \cite{ucf101}, EuroSAT \cite{eurosat}, Quick Draw \cite{QuickD}, Fungi \cite{Fungi}, Plant Disease \cite{PlantD} and Aircraft \cite{aircraft}. As demonstrated in our quantitative analysis in Section \ref{sec:mechanism}, specialized domains such as Fungi and Plant Disease introduce extreme textual domain shifts, offering a more representative evaluation of foundation model adaptation than standard ImageNet-centric suites.

% \subsubsection{\textbf{Datasets}} We choose datasets such that the sampled tasks are not too easy, cover different domains, and do not have many errors. In addition, in order to evaluate multimodal models, we require that each class of chosen datasets should have a text name. We finally choose ten datasets that satisfy these criteria: ImageNet-Sketch \cite{ImageNet_sketch}, DTD \cite{DTD}, CIFAR-100 \cite{CIFAR}, VGG Flowers \cite{VGG}, UCF-101 \cite{ucf101}, EuroSAT \cite{eurosat}, Quick Draw \cite{QuickD}, Fungi \cite{Fungi}, Plant Disease \cite{PlantD} and Aircraft \cite{aircraft}.
\color{black}
\subsubsection{\textbf{Base-novel split}} Following literature of transfer algorithms for multimodal models \cite{CoOP,Maple}, we split the classes of each dataset into a base set of classes and a novel set of classes. For base evaluation, the pretrained multimodal model will be adapted to the training set sampled from base set and evaluated on the test set sampled from base set. For base-to-novel evaluation, the pretrained multimodal model will still be adapted to the training set sampled from the base set, but evaluated on the test set sampled from the novel set of classes. This is possible since multimodal models like CLIP \cite{CLIP} do not need a tunable classification head, but classify images dependent on text names of classes only. For unimodal models, we only conduct base evaluation. The base-novel split is approximately $4:1$ for each dataset.

\subsubsection{\textbf{Sampling criteria}} We follow the task sampling criteria adopted in Meta-Dataset \cite{meta-dataset} with some small differences. Specifically, to sample a task, we first sample a random number of classes from the target task. The number of classes is sampled uniformly from $[2,15]$ for all datasets except for ImageNet-Sketch, whose classes per task are hierarchically sampled from one node in WordNet to improve the quality of sampled tasks. Then images in the task are sampled with an imbalance of shots for each class. In Meta-Dataset, the average number of shots can be large (20 or more), deviating from the true few-shot settings. We thus restrict the maximum number of training samples in each class to $10$, constructing ``true'' few-shot tasks. To have a well estimation of performance, we sample $600$ tasks per dataset and report the $95$\% confidence intervals. \textcolor{black}{To foster further progress in the field and facilitate reproducible research, we make the \FStransfer benchmark publicly available at \url{https://github.com/Frankluox/FewTrans}.}

\textcolor{black}{For the UCF-101 dataset, which is originally a video-based dataset, we adopt a frame-based classification protocol to evaluate few-shot transferability. Specifically, to minimize potential violations of statistical independence caused by temporal correlation, we strictly sample discrete frames from separate video clips to construct our tasks. This setup aligns with established practices in the multimodal transfer learning community, such as CoOp \cite{CoOP}, CoCoOp \cite{CoCoOp}, and MaPLe \cite{Maple}, ensuring a fair and direct comparison with current state-of-the-art baselines. While we acknowledge that intra-video correlations represent a structural characteristic of such data, our focus remains on recognizing semantic action categories across diverse classes.}

\subsection{Experiments on \FStransfer}
We use the aforementioned evaluation protocols to evaluate the few-shot transferability of pretrained models and compare different transfer algorithms. This results in three sub-benchmarks that (1) compares different pretrained models, (2) compares different transfer algorithms for pure vision models, and (3) compares different transfer algorithms for multimodal models for base evaluation and base-to-novel evaluation.

\subsubsection{\textbf{Evaluated models and algorithms}}  For pretrained models, we evaluate supervised models including ResNet-50 \cite{resnet}, SwinTransformer-base \cite{swin}, ConvNext-base \cite{convnext} trained on ImageNet 1K, and BiT-R101 \cite{BiT} trained on ImageNet 21K; self-supervised ImageNet models including MAE-base \cite{MAE}, IBOT-ViT-base \cite{IBOT} and EsViT-Swin-base \cite{EsViT}; multimodal pretrained model CLIP \cite{CLIP} trained on 400 millions image-text pairs; and self-supervised models DINOv2-small, DINOv2-base \cite{Dinov2} trained on 142M curated images. For transfer algorithms for pure vision models, we evaluate linear probing \cite{linear_probe}, Finetune \cite{MAE}, and several parameter-efficient finetuning methods including LoRA \cite{LoRA}, BitFit \cite{BitFit}, SSF \cite{SSF}, Adapter \cite{Adapter}, Adaptformer \cite{adaptformer}, VPT \cite{VPT} and TSA \cite{TSA}. For transfer algorithms for multimodal models, we evaluate CoOp \cite{CoOP}, CoCoOp \cite{CoCoOp}, VPT \cite{VPT}, MaPLe \cite{Maple}, KgCoOp \cite{kgcoop}, ProGrad \cite{ProGrad}, Finetune of visual encoder, Finetune of text encoder and Finetune of both encoders. We give results in Table \ref{benchmark:pretrained}-\ref{benchmark:transfermulmodalnovel}. We make following observations.

\subsection{Datasets}

\begin{enumerate}[leftmargin=*] 
% \item We observed  a different neural scaling law for the training of few-shot classification: test error falls off as a power law with the number of training classes, instead of the amount of training samples per class. 
    \item \textbf{ImageNet-Sketch} \cite{ImageNet_sketch} is a variant of ImageNet \cite{ImageNet} that contains 50000 sketches of all 1000 classes in ImageNet-1K. The
leaves of the wordnet that are reachable from `carnivore' and `device' form all 288 classes of novel set and other 712 classes belong to the base set.
\item \textbf{DTD} \cite{DTD} is a texture database containing  5640 classes with 47 classes. We randomly choose 37 classes as the base classes, and the other 10 classes as the novel classes.
\item \textbf{CIFAR-100} \cite{CIFAR} is a real-world dataset that has 100 classes containing 600 images each. We randomly choose 80 classes as the base classes, and the other 20 classes as the novel classes.

\item \textbf{VGG Flowers} \cite{VGG} is a dataset of natural images of 102 flower categories. Each category contains 40 to 258 images. We randomly choose 82 classes as the base classes, and the other 20 classes as the novel classes.

\item \textbf{UCF-101} \cite{ucf101} is an action recognition data set of realistic action videos, collected from YouTube, having 101 action categories with 13320 videos. We use sampled frames from these videos to form an image dataset.  We randomly choose 81 classes as the base classes, and the other 20 classes as the novel classes.

\item \textbf{EuroSAT} \cite{eurosat} is based on Sentinel-2 satellite images covering 13 spectral bands and consisting of 10 classes with in total 27000 labeled and geo-referenced images. We randomly choose 6 classes as the base classes, and the other 4 classes as the novel classes.

\item \textbf{Quick Draw} \cite{QuickD} is a dataset of black-and-white drawings across 345
categories. We use its smaller version used in DomainNet \cite{domainnet} that contains 172500 images. We randomly choose 276 classes as the base classes, and the other 69 classes as the novel classes. 

\item \textbf{Fungi} \cite{Fungi} is a dataset of 1394 categories of mashrooms species, with approximately 100K images. We randomly choose 1115 classes as the base classes, and the other 279 classes as the novel classes. 

\item \textbf{Plant Disease} \cite{PlantD} is a dataset that covers 38 categories of plant diseases. We randomly choose 30 classes as the base classes, and the other 8 classes as the novel classes. 

\item \textbf{Aircraft} \cite{aircraft} is a dataset containing 100 aircraft categories, with 100 images each. We randomly choose 80 classes as the base classes, and the other 20 classes as the novel classes.
\end{enumerate}

\subsection{Implementation Details}
To determine the range of hyperparameters for hyperparameter ensemble, we first need to find the average optimal hyperparameters on the held-out dataset. This procedure is exactly the same as the procedure we use in Section \ref{sec:Inappropriate}. Then for unimodal models that have two learning rates to be tuned, we expand the optimal hyperparameters into a $3\times3\times3$ grid. For the learning rate, we multiply the optimal hyperparameter found on the held-out dataset by 5 times and then divide it by 5 times to form the axis. For the number of epochs, we add 10 to the optimal hyperparameter, and expand it with its one-third and two-thirds to form the axis. For multimodal models and transfer algorithms that have only one learning rate to be tuned, we expand the optimal hyperparameters into a $5\times3$ grid, where the other rules remain unchanged. All points in the obtained grid will be used as hyperparameter configurations for the hyperparameter ensemble.

We conducted all experiments on 16 GeForce GTX 1080 Ti. For each evaluation of 600 tasks, the time cost on one GPU ranges from several hours to several days, depending on the pretrained models and transfer algorithms used. For all algorithms, we do not use weight decay for adaptation, and use default hyperparameters of Adam. For all transfer algorithms, we use their default settings of all other hyperparameters except for learning rates and the number of epochs. We set the batch size of the transfer process to be the maximum number that the GPU memory permits. During the evaluation, we fix the seed all the time, so in fact, the sequence of sampled tasks, and even the sequence of batch sampling inside each task, are exactly the same for all evaluation tasks, which ensures absolutely fair comparison throughout the process.

\begin{table*}[t]
% \scriptsize
\centering
\setlength\tabcolsep{1.6pt}
\caption{Sub-benchmark of \FStransfer that compares the few-shot transferability of different pretrained models. We use all-parameter finetune as the transfer algorithm for all models. We temporarily do not evaluate pretrained models that use larger architectures.}
\begin{tabular}{cc|cccccccccc|c}
% \hline
Models& Dataset & \rotatebox{90}{ImageNet-S} & \rotatebox{90}{DTD} & \rotatebox{90}{CIFAR-100} & \rotatebox{90}{Flowers} & \rotatebox{90}{UCF} & \rotatebox{90}{EuroSAT} & \rotatebox{90}{Quick Draw} & \rotatebox{90}{Fungi} & \rotatebox{90}{Plant Disease} & \rotatebox{90}{Aircraft} & \rotatebox{90}{Average}\\\hline
ResNet-50 & IN-1M& \acc{63.6}{1.5} & \acc{69.3}{1.1} & \acc{74.3}{1.1} & \acc{84.1}{1.1} & \acc{76.7}{1.2} & \acc{84.1}{1.0} & \acc{64.8}{1.3} & \acc{47.6}{1.4} & \acc{72.5}{1.4} & \acc{51.7}{1.4} & \acc{68.9}{1.3}\\
MAE-base & IN-1M& \acc{72.3}{1.5} & \acc{74.0}{1.1} & \acc{86.3}{0.8} & \acc{92.1}{0.6} & \acc{88.2}{0.8} & \acc{88.1}{0.9} & \acc{72.6}{1.2} & \acc{62.1}{1.3} & \acc{85.9}{0.8} & \acc{52.3}{1.3} & \acc{77.4}{1.1}\\
Swin-B & IN-1M& \acc{71.5}{1.5} & \acc{74.7}{1.1} & \acc{87.0}{0.8} & \acc{90.9}{0.8} & \acc{87.8}{0.8} & \acc{87.6}{0.9} & \acc{74.2}{1.1} & \acc{60.7}{1.3} & \acc{86.1}{0.8} & \acc{55.4}{1.4} & \acc{77.6}{1.1}\\
EsViT-SwinB & IN-1M& \acc{69.7}{1.5} & \acc{75.3}{1.0} & \acc{84.7}{0.9} & \acc{93.6}{0.6} & \acc{88.0}{0.8} & \acc{87.8}{0.9} & \acc{70.8}{1.2} & \acc{62.9}{1.3} & \acc{86.2}{0.8} & \acc{57.7}{1.5} & \acc{77.7}{1.1}\\
ConvNext-B& IN-1M& \acc{74.6}{1.5} & \acc{73.9}{1.1} & \acc{88.3}{0.7} & \acc{90.8}{0.8} & \acc{89.1}{0.8} & \acc{86.4}{1.0} & \acc{75.1}{1.1} & \acc{60.8}{1.3} & \acc{84.2}{0.9} & \acc{55.7}{1.4} & \acc{77.9}{1.1}\\
IBOT-ViT & IN-1M& \acc{69.7}{1.5} & \acc{74.9}{1.1} & \acc{89.6}{0.7} & \acc{93.6}{0.6} & \acc{89.1}{0.8} & \acc{\textbf{89.1}}{0.8} & \acc{69.9}{1.2} & \acc{60.8}{1.3} & \acc{86.3}{0.8} & \acc{56.7}{1.5} & \acc{78.0}{1.1}\\
BiT-R101 & IN-14M& \acc{68.2}{1.5} & \acc{77.3}{1.0} & \acc{85.3}{0.8} & \acc{99.6}{0.1} & \acc{89.4}{0.7} & \acc{87.0}{0.9} & \acc{71.9}{1.2} & \acc{67.9}{1.2} & \acc{91.1}{0.6} & \acc{57.5}{1.4} & \acc{79.5}{1.0}\\
CLIP-base & WIT-400M& \acc{\textbf{80.4}}{1.3} & \acc{\textbf{83.5}}{0.8} & \acc{\textbf{94.9}}{0.3} & \acc{96.9}{0.3} & \acc{\textbf{95.4}}{0.3} & \acc{88.5}{0.7} & \acc{76.5}{0.9} & \acc{54.8}{1.4} & \acc{77.2}{1.0} & \acc{\textbf{78.8}}{1.0} & \acc{82.7}{0.9}\\
DINOv2-small &LVD-142M& \acc{75.1}{1.4} & \acc{81.3}{0.9} & \acc{89.8}{0.7} & \acc{99.6}{0.1} & \acc{90.3}{0.7} & \acc{87.0}{0.9} & \acc{78.1}{1.0} & \acc{69.9}{1.2} & \acc{89.8}{0.7} & \acc{67.3}{1.4} & \acc{82.8}{1.0}\\
DINOv2-base &LVD-142M& \acc{79.8}{1.3} & \acc{82.6}{0.9} & \acc{93.0}{0.5} & \acc{\textbf{99.9}}{0.0} & \acc{93.9}{0.5} & \acc{87.7}{0.9} & \acc{\textbf{79.6}}{1.0} & \acc{\textbf{74.9}}{1.1} & \acc{\textbf{91.8}}{0.6} & \acc{70.3}{1.3} & \acc{\textbf{85.3}}{0.9}\\

\hline
\end{tabular}
\label{benchmark:pretrained}
\vspace{-0in}
\end{table*}

\begin{table*}[t]
% \footnotesize
% \scriptsize
\centering
\setlength\tabcolsep{3pt}
\caption{Sub-benchmark of \FStransfer that compares 
different transfer algorithms for pure vision pretrained models. The visual encoder of CLIP-base is chosen as the pretrained model.}
\begin{tabular}{c|cccccccccc|c}
% \hline
& \rotatebox{90}{ImageNet-S} & \rotatebox{90}{DTD} & \rotatebox{90}{CIFAR-100} & \rotatebox{90}{Flowers} & \rotatebox{90}{UCF} & \rotatebox{90}{EuroSAT} & \rotatebox{90}{Quick Draw} & \rotatebox{90}{Fungi} & \rotatebox{90}{Plant Disease} & \rotatebox{90}{Aircraft} & \rotatebox{90}{Average}\\\hline

\hline
Linear & \acc{72.1}{1.5} & \acc{76.7}{1.1} & \acc{83.7}{0.8} & \acc{95.5}{0.4} & \acc{91.6}{0.7} & \acc{81.5}{1.0} & \acc{70.8}{1.1} & \acc{56.8}{1.4} & \acc{75.3}{1.1} & \acc{68.0}{1.3} & \acc{77.2}{1.1}\\
Finetune & \acc{73.1}{1.5} & \acc{79.9}{1.0} & \acc{88.0}{0.8} & \acc{95.9}{0.5} & \acc{93.0}{0.6} & \acc{87.6}{0.9} & \acc{\textbf{78.9}}{1.0} & \acc{58.9}{1.4} & \acc{83.7}{0.9} & \acc{70.7}{1.3} & \acc{81.0}{1.0}\\
LoRA & \acc{73.8}{1.5} & \acc{80.7}{1.0} & \acc{88.7}{0.7} & \acc{96.1}{0.4} & \acc{93.3}{0.6} & \acc{\textbf{87.7}}{0.9} & \acc{78.0}{1.1} & \acc{59.4}{1.4} & \acc{83.3}{0.9} & \acc{71.0}{1.3} & \acc{81.2}{1.0}\\
BitFit & \acc{73.6}{1.5} & \acc{79.7}{1.0} & \acc{89.3}{0.7} & \acc{96.8}{0.4} & \acc{93.3}{0.6} & \acc{86.5}{0.9} & \acc{77.3}{1.1} & \acc{61.3}{1.3} & \acc{83.5}{0.9} & \acc{71.0}{1.2} & \acc{81.2}{1.0}\\
SSF & \acc{74.2}{1.5} & \acc{80.3}{1.0} & \acc{89.0}{0.7} & \acc{96.7}{0.4} & \acc{93.2}{0.6} & \acc{87.4}{0.9} & \acc{77.3}{1.1} & \acc{60.8}{1.3} & \acc{84.4}{0.9} & \acc{70.7}{1.3} & \acc{81.4}{1.0}\\
Adapter & \acc{74.1}{1.5} & \acc{80.5}{1.0} & \acc{89.8}{0.7} & \acc{96.9}{0.4} & \acc{93.6}{0.5} & \acc{86.5}{0.9} & \acc{77.3}{1.0} & \acc{61.2}{1.3} & \acc{83.2}{0.9} & \acc{70.9}{1.2} & \acc{81.4}{1.0}\\
Adaptformer & \acc{74.1}{1.5} & \acc{80.8}{1.0} & \acc{90.0}{0.7} & \acc{\textbf{97.0}}{0.3} & \acc{\textbf{93.8}}{0.5} & \acc{87.0}{0.9} & \acc{77.7}{1.0} & \acc{61.8}{1.3} & \acc{83.6}{0.9} & \acc{71.0}{1.2} & \acc{81.7}{1.0}\\
VPT & \acc{73.2}{1.5} & \acc{\textbf{82.1}}{0.9} & \acc{\textbf{90.2}}{0.7} & \acc{\textbf{97.0}}{0.4} & \acc{93.6}{0.5} & \acc{87.3}{0.9} & \acc{78.2}{1.0} & \acc{\textbf{61.9}}{1.3} & \acc{85.7}{0.9} & \acc{71.6}{1.2} & \acc{82.1}{1.0}\\
TSA & \acc{\textbf{74.3}}{1.5} & \acc{80.0}{1.0} & \acc{89.5}{0.7} & \acc{96.9}{0.4} & \acc{93.5}{0.6} & \acc{87.5}{0.9} & \acc{78.3}{1.0} & \acc{64.5}{1.3} & \acc{\textbf{86.2}}{0.8} & \acc{\textbf{72.2}}{1.2} & \acc{\textbf{82.3}}{1.0}\\

\hline

\end{tabular}
\label{benchmark:transfervision}
\end{table*}
%base

\subsection{Main Results and Analysis}
\subsubsection{\textbf{The size of the pretraining dataset matters}} As seen from Table \ref{benchmark:pretrained}, models trained on ImageNet-1K have very similar performance when well-tuned (except for ResNet-50 which does not use most of the training tricks), regardless of the training algorithm and architecture used. The difference between the worst-performing MAE and best-performing IBOT is $0.6$, smaller than the range of confidence interval. However, when the dataset size increases, we see a very clear improvement in few-shot transfer performance.

\begin{table*}[t]
% \footnotesize
% \scriptsize
\centering
\setlength\tabcolsep{2pt}
\caption{Sub-benchmark of \FStransfer that compares 
different transfer algorithms for base evaluation of multi-modal pretrained models. CLIP-base is chosen as the pretrained model.}
\begin{tabular}{c|cccccccccc|c}
% \hline
& \rotatebox{90}{ImageNet-S} & \rotatebox{90}{DTD} & \rotatebox{90}{CIFAR-100} & \rotatebox{90}{Flowers} & \rotatebox{90}{UCF} & \rotatebox{90}{EuroSAT} & \rotatebox{90}{Quick Draw} & \rotatebox{90}{Fungi} & \rotatebox{90}{Plant Disease} & \rotatebox{90}{Aircraft} & \rotatebox{90}{Average}\\\hline
\textcolor[RGB]{128,128,128}{
Zero-shot} & \acc{72.6}{1.5} & \acc{73.0}{1.0} & \acc{92.9}{0.4} & \acc{86.3}{0.9} & \acc{90.5}{0.6} & \acc{64.4}{1.2} & \acc{57.4}{1.2} & \acc{38.7}{1.5} & \acc{46.0}{1.4} & \acc{69.2}{1.2} & \acc{69.1}{1.2}\\
CoOp & \acc{79.3}{1.3} & \acc{83.8}{0.8} & \acc{93.8}{0.4} & \acc{97.8}{0.2} & \acc{95.1}{0.4} & \acc{84.3}{0.8} & \acc{73.8}{0.9} & \acc{51.9}{1.5} & \acc{70.9}{1.2} & \acc{70.0}{1.4} & \acc{80.1}{1.0}\\
ProGrad & \acc{79.4}{1.3} & \acc{82.3}{0.8} & \acc{93.9}{0.4} & \acc{96.2}{0.3} & \acc{94.7}{0.4} & \acc{84.1}{0.8} & \acc{72.5}{0.9} & \acc{53.8}{1.4} & \acc{71.6}{1.1} & \acc{73.2}{1.2} & \acc{80.2}{0.9}\\
VPT & \acc{78.8}{1.3} & \acc{81.3}{0.8} & \acc{94.5}{0.3} & \acc{95.5}{0.4} & \acc{94.5}{0.4} & \acc{88.3}{0.7} & \acc{75.1}{0.9} & \acc{47.4}{1.5} & \acc{72.9}{1.1} & \acc{76.5}{1.1} & \acc{80.5}{0.9}\\
MaPLe & \acc{79.2}{1.3} & \acc{82.5}{0.8} & \acc{94.6}{0.3} & \acc{96.5}{0.4} & \acc{95.1}{0.4} & \acc{88.8}{0.7} & \acc{76.3}{0.9} & \acc{48.9}{1.5} & \acc{74.6}{1.1} & \acc{74.5}{1.1} & \acc{81.1}{0.9}\\
KgCoOp & \acc{79.9}{1.2} & \acc{84.1}{0.7} & \acc{94.1}{0.4} & \acc{97.5}{0.2} & \acc{95.3}{0.4} & \acc{84.7}{0.8} & \acc{74.1}{0.9} & \acc{55.2}{1.5} & \acc{72.9}{1.1} & \acc{73.9}{1.2} & \acc{81.2}{0.9}\\
CoCoOp & \acc{79.8}{1.2} & \acc{83.4}{0.8} & \acc{93.8}{0.4} & \acc{97.4}{0.3} & \acc{95.4}{0.4} & \acc{86.3}{0.7} & \acc{76.0}{0.9} & \acc{52.2}{1.6} & \acc{76.7}{1.1} & \acc{74.1}{1.2} & \acc{81.5}{1.0}\\
AllFT & \acc{80.4}{1.3} & \acc{83.5}{0.8} & \acc{94.9}{0.3} & \acc{96.9}{0.3} & \acc{95.4}{0.3} & \acc{88.5}{0.7} & \acc{76.5}{0.9} & \acc{54.8}{1.4} & \acc{77.2}{1.0} & \acc{78.8}{1.0} & \acc{82.7}{0.9}\\
VisualFT & \acc{80.0}{1.2} & \acc{83.0}{0.8} & \acc{\textbf{95.1}}{0.3} & \acc{96.6}{0.4} & \acc{95.1}{0.4} & \acc{\textbf{89.9}}{0.7} & \acc{\textbf{78.3}}{0.8} & \acc{52.7}{1.4} & \acc{80.1}{0.9} & \acc{77.7}{1.0} & \acc{82.9}{0.9}\\
TextFT & \acc{\textbf{80.9}}{1.2} & \acc{\textbf{85.4}}{0.7} & \acc{94.2}{0.4} & \acc{\textbf{98.3}}{0.2} & \acc{\textbf{96.0}}{0.3} & \acc{85.6}{0.8} & \acc{75.8}{0.9} & \acc{\textbf{62.5}}{1.4} & \acc{\textbf{80.3}}{0.9} & \acc{\textbf{79.0}}{1.0} & \acc{83.8}{0.9}\\

\hline
\end{tabular}
\label{benchmark:transfermulmodalbase}
\end{table*}

\begin{table*}[t]
% \footnotesize
% \scriptsize
\centering
\setlength\tabcolsep{2pt}
\caption{Sub-benchmark of \FStransfer that compares 
different transfer algorithms for base-to-novel evaluation of multi-modal pretrained models. CLIP-base is chosen as the pretrained model.}
\begin{tabular}{c|cccccccccc|c}
% \hline
& \rotatebox{90}{ImageNet-S} & \rotatebox{90}{DTD} & \rotatebox{90}{CIFAR-100} & \rotatebox{90}{Flowers} & \rotatebox{90}{UCF} & \rotatebox{90}{EuroSAT} & \rotatebox{90}{Quick Draw} & \rotatebox{90}{Fungi} & \rotatebox{90}{Plant Disease} & \rotatebox{90}{Aircraft} & \rotatebox{90}{Average}\\\hline

\hline
CoCoOp & \acc{67.5}{1.4} & \acc{66.7}{1.1} & \acc{86.8}{0.5} & \acc{77.5}{1.1} & \acc{87.9}{0.7} & \acc{67.5}{1.3} & \acc{61.4}{1.1} & \acc{21.8}{1.2} & \acc{59.4}{1.4} & \acc{47.9}{1.7} & \acc{64.4}{1.2}\\
CoOp & \acc{64.3}{1.5} & \acc{71.5}{1.0} & \acc{86.5}{0.6} & \acc{85.0}{0.8} & \acc{86.1}{0.7} & \acc{71.5}{1.2} & \acc{\textbf{60.9}}{1.2} & \acc{31.5}{1.4} & \acc{65.0}{1.2} & \acc{46.2}{1.9} & \acc{66.8}{1.2}\\
ProGrad & \acc{65.0}{1.5} & \acc{71.7}{1.0} & \acc{86.7}{0.5} & \acc{85.0}{0.8} & \acc{86.3}{0.7} & \acc{72.0}{1.2} & \acc{61.1}{1.2} & \acc{32.5}{1.4} & \acc{65.7}{1.2} & \acc{49.7}{1.8} & \acc{67.6}{1.2}\\
VPT & \acc{71.7}{1.3} & \acc{67.7}{1.0} & \acc{\textbf{87.5}}{0.6} & \acc{84.5}{0.8} & \acc{86.4}{0.7} & \acc{68.1}{1.4} & \acc{56.7}{1.2} & \acc{37.0}{1.3} & \acc{56.9}{1.4} & \acc{61.5}{1.3} & \acc{67.8}{1.1}\\
MaPLe & \acc{70.4}{1.3} & \acc{62.1}{1.2} & \acc{88.3}{0.5} & \acc{82.4}{0.8} & \acc{87.3}{0.6} & \acc{\textbf{77.1}}{1.3} & \acc{60.8}{1.1} & \acc{34.3}{1.3} & \acc{62.2}{1.3} & \acc{56.2}{1.4} & \acc{68.1}{1.1}\\
KgCoOp & \acc{68.9}{1.4} & \acc{\textbf{72.7}}{0.9} & \acc{87.0}{0.5} & \acc{86.6}{0.7} & \acc{87.8}{0.7} & \acc{70.6}{1.2} & \acc{60.6}{1.1} & \acc{33.9}{1.4} & \acc{66.7}{1.2} & \acc{51.9}{1.8} & \acc{68.7}{1.2}\\
\textcolor[RGB]{128,128,128}{
Zero-shot} & \acc{73.9}{1.3} & \acc{68.7}{1.1} & \acc{86.8}{0.5} & \acc{87.0}{0.7} & \acc{89.0}{0.6} & \acc{69.7}{1.4} & \acc{58.1}{1.2} & \acc{39.3}{1.4} & \acc{59.2}{1.2} & \acc{61.5}{1.3} & \acc{69.3}{1.1}\\
VisualFT & \acc{74.0}{1.3} & \acc{69.0}{1.0} & \acc{88.3}{0.5} & \acc{86.7}{0.7} & \acc{89.0}{0.6} & \acc{70.2}{1.4} & \acc{60.4}{1.1} & \acc{38.9}{1.4} & \acc{67.5}{1.3} & \acc{62.2}{1.3} & \acc{70.6}{1.1}\\
TextFT & \acc{\textbf{74.2}}{1.3} & \acc{69.8}{1.0} & \acc{87.0}{0.5} & \acc{\textbf{87.5}}{0.7} & \acc{89.8}{0.6} & \acc{72.2}{1.4} & \acc{59.9}{1.2} & \acc{39.2}{1.4} & \acc{\textbf{70.2}}{1.2} & \acc{61.7}{1.3} & \acc{71.2}{1.1}\\
AllFT & \acc{74.1}{1.3} & \acc{69.4}{1.0} & \acc{88.1}{0.5} & \acc{87.2}{0.7} & \acc{\textbf{89.5}}{0.6} & \acc{72.3}{1.4} & \acc{\textbf{60.9}}{1.1} & \acc{\textbf{39.6}}{1.4} & \acc{68.9}{1.2} & \acc{\textbf{62.8}}{1.3} & \acc{\textbf{71.3}}{1.1}\\

\hline

\end{tabular}
\label{benchmark:transfermulmodalnovel}
\end{table*}

\subsubsection{\textbf{CLIP meets problems with uncommon class names}} From Figure \ref{benchmark:pretrained}, we see that CLIP exhibits promising performance on most datasets, but performs badly on Fungi and Plant Disease, two fine-grained datasets whose category names are mostly rare words. This is something like a ``text domain shift'' which requires significant updates for the text encoder. We expect that such problems can be relieved when the number of shots increases, but for few-shot evluation on these two datasets, only using the visual encoder of CLIP (see Table \ref{benchmark:transfervision}) can be better than using both encoders (see Table \ref{benchmark:transfermulmodalbase}).

\subsubsection{\textbf{Visual-only transfer algorithms perform similar}} From Table \ref{benchmark:transfervision}, we can see that except for linear probing, all transfer algorithms for pure visual pretrained models have very similar performance and have intersected confidence intervals. This is in contrast to the benchmark of many-shot transfer learning like VTAB \cite{VTAB}, where different transfer algorithms are shown to have significant performance gaps (see \cite{chavan2023one} for example). 

\subsubsection{\textbf{Finetune performs surprisingly well}} on all sub-benchmarks for transfer algorithms as shown in Table \ref{benchmark:transfervision}-\ref{benchmark:transfermulmodalnovel}, especially for multimodal models. Intuitively, finetuning all parameters of the pretrained model with a few samples is expected to encounter significant overfitting issues. Such a phenomenon needs deeper understanding.

\subsubsection{\textbf{Are we making progress on few-shot multimodal transfer?}} While we observe in Table \ref{benchmark:transfermulmodalbase} that all specifically designed transfer algorithms for CLIP perform better than zero-shot baseline in base evaluation, they all perform worse than zero-shot baseline in base-to-novel evaluation in Table \ref{benchmark:transfermulmodalnovel}, different from what some of the methods claimed in their paper with the old benchmarks. In contrast, simple finetune, either finetuning a single encoder or finetuning both, surpasses all these methods in both evaluation settings. This indicates we are \emph{not} making progress in this field and we should rethink what’s the thing that leads to real improvement of few-shot multimodal transfer performance.
\color{black}
\subsection{Statistical Significance and Resource Disclosure}
\label{sec:stats_resource}

\keypoint{Statistical significance analysis.} To provide a rigorous foundation for our observation that transfer algorithms perform similarly in few-shot regimes, we conducted a paired statistical analysis between full fine-tuning (Full-FT) and LoRA across 6,000 tasks (600 tasks $\times$ 10 datasets). As summarized in Table \ref{tab:statistical_tests}, while the large aggregate sample size leads to a low global p-value ($9\times10^{-7}$), the overall Cohen's $d$ effect size is only $-0.22$. In statistical theory, an effect size of approximately $0.2$ is classified as ``small'' or ``negligible,'' confirming that the practical difference between these methods is minimal. Specifically, on 6 out of 10 datasets (e.g., Aircraft, Fungi, Quick Draw), no statistically significant difference was detected ($p > 0.05$). Even where differences were detected, the absolute accuracy gap remained narrow, typically within 1\%--2\%.

\begin{table}[t]
\centering
\color{black}
\caption{Paired t-test results and Cohen's $d$ effect sizes between Full-FT and LoRA ($N=600$ tasks per dataset).}
\label{tab:statistical_tests}
\setlength\tabcolsep{3pt}
\begin{tabular}{l|ccccc}
\hline
Dataset & Full-FT (\%) & LoRA (\%) & Diff (\%) & $p$-value & Cohen's $d$ \\ \hline
Aircraft & 63.39 & 63.81 & -0.42 & 0.4859 & -0.10 \\
CIFAR100 & 82.21 & 84.45 & -2.24 & 0.0003 & -0.55 \\
Fungi & 55.73 & 55.84 & -0.11 & 0.8646 & -0.02 \\
ILSVRC & 85.49 & 87.17 & -1.68 & 0.0004 & -0.54 \\
Quick Draw & 72.45 & 73.23 & -0.78 & 0.1794 & -0.19 \\
Textures & 68.69 & 69.76 & -1.07 & 0.0562 & -0.28 \\
VGG Flower & 96.21 & 97.20 & -0.99 & 0.0012 & -0.49 \\
EuroSAT & 69.15 & 70.96 & -1.81 & 0.0092 & -0.38 \\
Plant Disease & 83.71 & 82.88 & 0.83 & 0.1102 & 0.23 \\
UCF101 & 91.07 & 91.31 & -0.24 & 0.5182 & -0.09 \\ \hline
Overall & 76.81 & 77.66 & -0.85 & $9\times10^{-7}$ & -0.22 \\ \hline
\end{tabular}
\end{table}

\keypoint{Resource disclosure and feasibility.} We report the computational requirements for evaluating 600 tasks on a single NVIDIA TITAN Xp GPU in Table \ref{tab:computational_cost}. \textcolor{black}{For users with limited hardware (e.g., a single consumer-grade GPU), HPE remains highly accessible via serial execution, as it does not increase the peak GPU memory requirement ($\approx 12.1$ GB for Full-FT).}  Under this serial execution, full fine-tuning, representing the upper bound of adaptation cost, requires approximately 13.3 hours and 12.1 GB of peak memory. Parameter-efficient methods like LoRA significantly reduce this to 5.5 hours and 7.1 GB. 
\textcolor{black}{Importantly, since each configuration in the hyperparameter grid is completely independent, they can also be evaluated simultaneously across multiple GPUs or workers. On modern multi-GPU setups, the wall-clock time for an entire evaluation session can be drastically reduced, approaching the time of a single fine-tuning run ($T_{HPE} \approx T_{single}$). For larger search spaces, strategies such as random-grid sampling can be employed to maintain tractability. These results demonstrate that the \FStransfer benchmark is highly feasible for academic research using consumer-grade hardware.}

\begin{table}[t]
\centering
\color{black}
\caption{Computational requirements for evaluating 600 tasks on a single GPU.}
\label{tab:computational_cost}
\begin{tabular}{l|cc}
\hline
Method & Total Time (600 Tasks) & Peak GPU Memory \\ \hline
Full Fine-tuning & 13.3 hours & $\approx$12,100 MB \\
LoRA & 5.5 hours & $\approx$7,100 MB \\ \hline
\end{tabular}
\end{table}

\color{black}
\section{Mechanism Analysis of Fine-tuning}
\label{sec:mechanism}

To address the counter-intuitive observation that full-parameter fine-tuning (Full-FT) consistently outperforms parameter-efficient fine-tuning (PEFT) methods without overfitting in extreme few-shot regimes, we conduct a deeper investigation into the adaptation mechanism from two perspectives: parameter update scales and feature distribution shifts.

\begin{figure*}[t]
    \centering
    \includegraphics[width=0.8\linewidth]{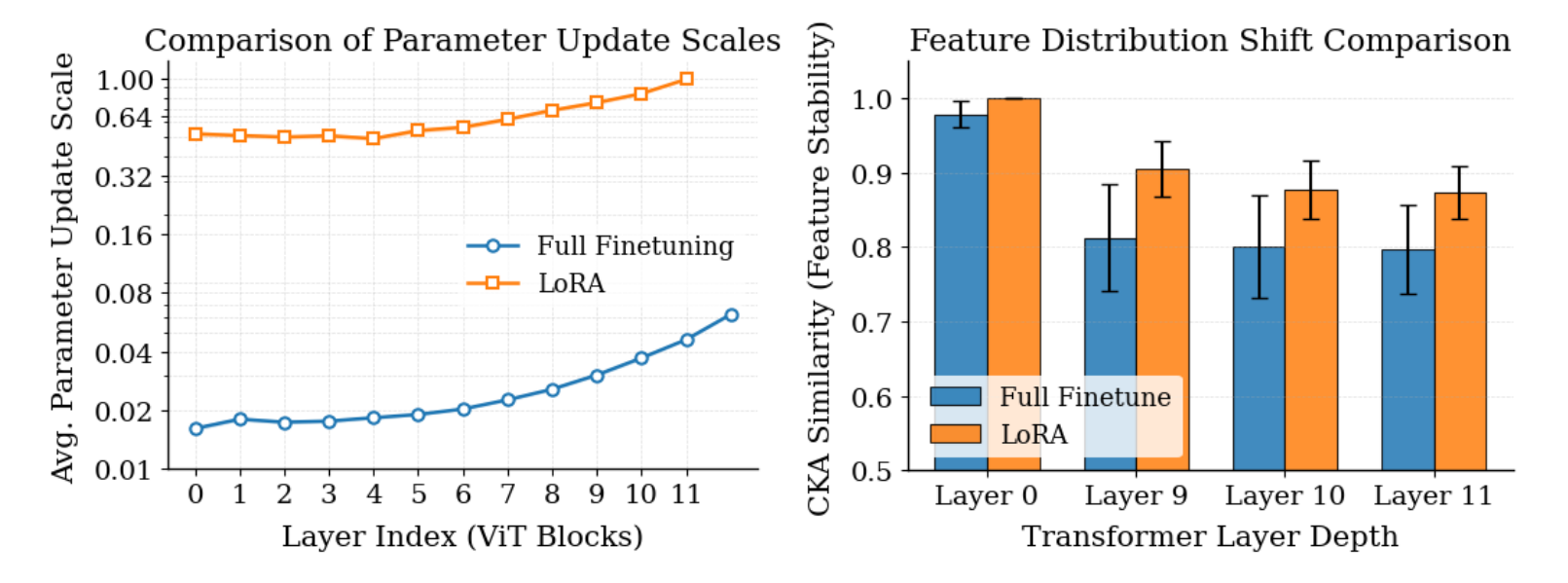}
    \color{black}
    \caption{Analysis of the adaptation mechanism for full fine-tuning and LoRA. (Left) Layer-wise parameter update scales ($L_{2}$ norm of $\Delta W$ across 12 Transformer blocks, showing that Full fine-tuning relies on distributed micro-adjustments ($0.01 \sim 0.07$) to avoid overfitting. (Right) Feature distribution shift measured by Centered Kernel Alignment (CKA) similarity at deep layers, where Full fine-tuning demonstrates more effective reshaping of high-level semantic representations compared to the constrained adaptation of LoRA.}
    \label{fig:mechanism_analysis}
\end{figure*}

\subsection{Parameter Update Scales: The Micro-adjustment Principle}
We analyze the layer-wise parameter update scales, defined as the $L_{2}$ norm of the weight difference $\Delta W = W_{adapted} - W_{initial}$. As shown in Figure \ref{fig:mechanism_analysis}, two critical patterns emerge:

\keypoint{Distributed micro-adjustments.} Full-FT achieves superior performance with significantly smaller update magnitudes, typically ranging from $0.01$ to $0.07$, compared to LoRA, which exhibits shifts between $0.4$ and $1.0$. This indicates that Full-FT adapts to downstream tasks through subtle ``micro-adjustments'' distributed across the entire parameter space. The proximity to pre-trained initialization serves as an implicit regularization, providing a principled explanation for why Full-FT avoids the expected overfitting in extreme few-shot regimes.

\keypoint{Back-heavy update pattern.} Both Full-FT and LoRA exhibit a ``back-heavy'' update pattern: earlier layers (Blocks 0-4) remain relatively stable to preserve foundational features, while deeper blocks (Blocks 8-11) undergo more substantial modifications. Crucially, Full-FT exhibits a sharper relative incline in final blocks compared to LoRA, suggesting it is more effective at ``unlocking'' and re-specializing high-level semantic representations to the target domain.

\textcolor{black}{\textbf{Connecting update scales to landscape flatness.} Our empirical analysis reveals that Full Fine-tuning (Full-FT) results in an extremely small parameter displacement, with $L_2$ norms ranging from $0.01$ to $0.07$ across Transformer blocks. This behavior acts as a form of \emph{implicit regularization}. Specifically, since large-scale pre-trained models like DINOv2 typically converge into broad, flat minima basins that exhibit superior generalization, these distributed micro-adjustments ensure that the adapted model parameters remain within the \emph{immediate local basin} of the pre-trained weights. By contrast, larger parameter shifts (as seen in PEFT methods like LoRA) may force the model to escape this flat region and converge into ``sharp'' local minima that are more prone to overfitting on scarce support samples. Thus, Full-FT achieves robust transfer by inheriting the pre-trained landscape flatness while performing subtle task specialization.}

\subsection{Feature Distribution Shift via CKA}
To complement the parameter-level analysis, we further investigate feature-level behavior using Centered Kernel Alignment (CKA) to measure the similarity between pre-trained and adapted feature distributions.

\keypoint{Effective reshaping of deep features.} As shown in Figure \ref{fig:mechanism_analysis}, Full-FT exhibits significantly lower CKA similarity in deeper layers (Layers 9-11) compared to LoRA. This lower similarity indicates a greater feature distribution shift, demonstrating that Full-FT more flexibly reshapes high-level semantic representations to align with the target task. In contrast, the feature adaptation of LoRA is notably more constrained.

\keypoint{Stability and synergy.} Both methods show high CKA similarity in the initial layer (Layer 0), indicating foundational features are preserved. The synergy between subtle parameter updates and substantial feature-level shifts in deep layers allows Full-FT to achieve a more thorough adaptation to downstream task distributions than parameter-efficient alternatives.

\begin{figure*}[t]
    \centering
    \includegraphics[width=0.7\linewidth]{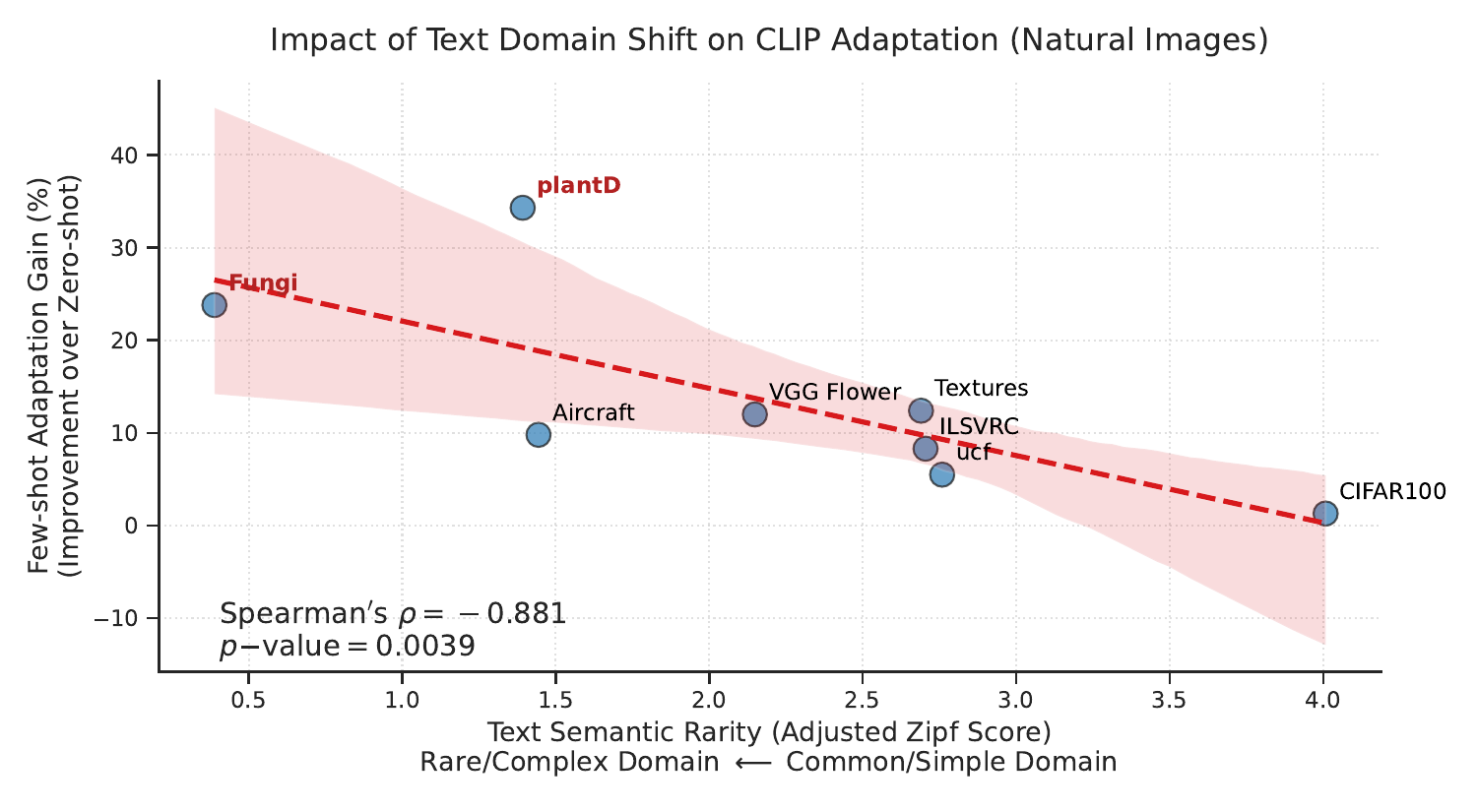}
    \color{black}
    \caption{Quantitative analysis of text domain shift on CLIP few-shot adaptation. The scatter plot illustrates a strong negative Spearman rank correlation ($\rho=-0.881, p=0.0039$) between text semantic rarity (adjusted Zipf score) and few-shot adaptation gain. Datasets with rarer linguistic terms (e.g., Fungi, plantD) exhibit significant zero-shot performance collapses, necessitating indispensable correction via full fine-tuning to realign visual features.}
    \label{fig:text_domain_shift}
\end{figure*}

\subsection{Characterizing and Quantifying Text Domain Shift}
The performance collapse of multimodal models on specific datasets is quantitatively addressed using adjusted Zipf frequency scores. We identify three primary dimensions of this shift:
\begin{itemize}
    \item \textbf{Vocabulary shift}: In the \textit{Fungi} dataset, category names are dominated by scientific Latin terms (e.g., \textit{Achroomyces disciformis} and \textit{Agaricus cupreobrunneus}). These terms are nearly absent in CLIP’s general-purpose pre-training corpus, rendering the text encoder's initial semantic anchors ineffective.
    \item \textbf{Compositional semantic shift}: In the \textit{Plant Disease (plantD)} dataset, while individual tokens like ``leaf'' or ``apple'' are frequent, their specific pathological combinations (e.g., \textit{cedar apple rust leaf} or \textit{gray leaf spot of corn leaf}) create a semantic bottleneck due to their rarity in natural language descriptions.
    \item \textbf{Technical and alphanumeric identifiers}: In datasets like \textit{Aircraft}, labels often revert to abstract alphanumeric strings (e.g., \textit{707-320}, \textit{A340-600}, or \textit{BAE 146-200}). Such identifiers lack inherent natural language context, forcing the model to rely on Full-FT to bridge the gap between visual features and these arbitrary symbolic labels.
\end{itemize}

% This metric captures vocabulary shifts (e.g., rare Latin terminology in Fungi) and compositional semantic shifts (e.g., specific pathological combinations in Plant Disease).

As shown in Figure \ref{fig:text_domain_shift}, we observe a strong negative Spearman rank correlation ($\rho=-0.881, p=0.0039$) between text semantic rarity and few-shot adaptation gain. This relationship indicates that when the text encoder fails to provide meaningful semantic anchors due to domain shift, Full-FT becomes an indispensable correction mechanism. It compensates for the ``textual silence'' by re-aligning visual features directly from few-shot labels, bridging the gap that zero-shot or prompt-based methods cannot fill.

\color{black}
% \section{Conclusion and Future Work}
% We have introduced \FStransfer, a unified, realistic, rigorous benchmark for evaluating few-shot transferability of pretrained models. Our initial exploration of this benchmark shows that transferring from a better pretrained model trained on a large pretraining dataset seems to be much more important than using a better transfer algorithm. However, we believe that with rigorous evaluation, comparison and further investigations on \FStransfer, good transfer algorithms will finally emerge. We are now implementing more algorithms and trying to include more pretrained models in the benchmark. In addition to comparing few-shot performance, we plan to add a comparison of the number of tunable parameters and the time needed for a complete adaptation for transfer algorithms.

\section{Conclusion and future work}
\label{sec:conclusion}

We have introduced \FStransfer, a unified, realistic, and rigorous benchmark designed to evaluate the few-shot transferability of pre-trained models. Our extensive empirical study reveals that the choice of the pre-trained model architecture and its training data scale are significantly more dominant factors for performance than the transfer algorithm itself. While simple fine-tuning remains a surprisingly strong baseline, we believe that the established standards of \FStransfer will drive the community toward developing truly innovative adaptation methods that move beyond the ``validation set illusion''.

\color{black}
\keypoint{Limitations and extensibility.} While \FStransfer addresses critical evaluation flaws, we acknowledge its current focus on standard task accuracy. There is significant potential to extend this benchmark into a multi-dimensional diagnostic tool. First, the protocol can be adapted to evaluate model robustness against adversarial attacks, ensuring that transferred models remain reliable under malicious perturbations in security-sensitive scenarios. Second, we identify out-of-distribution (OOD) generalization within few-shot tasks as a vital research direction. Future iterations of the benchmark could evaluate how well a model adapted on a tiny support set generalizes to test sets with significant domain shifts or covariate shifts. By addressing these dimensions, we aim to provide a more comprehensive ``ruler'' for the foundation model era.

\color{black}

% \section{Reproducibility Statemment}
% We do our best to ensure the reproducibility of our benchmark. We include most  details of our empirical investigations and the benchmark in the two sections of Appendix. The code for the benchmark can be found at \url{https://anonymous.4open.science/r/FewTrans-7FB5}.

\section*{Acknowledgments}
This study is supported by grants from the National Natural Science Foundation of China (No.U22A2097, No.62506310, No.U23A20315, No.82441006, No.62425208).

\bibliographystyle{IEEEtran}
\bibliography{ref}

@inproceedings{BiT,
  title={Big transfer (bit): General visual representation learning},
  author={Kolesnikov, Alexander and Beyer, Lucas and Zhai, Xiaohua and Puigcerver, Joan and Yung, Jessica and Gelly, Sylvain and Houlsby, Neil},
  booktitle={ECCV},
  year={2020},
}

@inproceedings{CLIP,
  title={Learning transferable visual models from natural language supervision},
  author={Radford, Alec and Kim, Jong Wook and Hallacy, Chris and Ramesh, Aditya and Goh, Gabriel and Agarwal, Sandhini and Sastry, Girish and Askell, Amanda and Mishkin, Pamela and Clark, Jack and others},
  booktitle={ICML},
  year={2021},
}

@article{CoOP,
  title={Learning to prompt for vision-language models},
  author={Zhou, Kaiyang and Yang, Jingkang and Loy, Chen Change and Liu, Ziwei},
  journal={IJCV},
  year={2022},
}

@article{Dinov2,
  title={Dinov2: Learning robust visual features without supervision},
  author={Oquab, Maxime and Darcet, Timoth{\'e}e and Moutakanni, Th{\'e}o and Vo, Huy and Szafraniec, Marc and Khalidov, Vasil and Fernandez, Pierre and Haziza, Daniel and Massa, Francisco and El-Nouby, Alaaeldin and others},
  journal={arXiv preprint arXiv:2304.07193},
  year={2023}
}

@inproceedings{dehghani2023scaling,
  title={Scaling vision transformers to 22 billion parameters},
  author={Dehghani, Mostafa and Djolonga, Josip and Mustafa, Basil and Padlewski, Piotr and Heek, Jonathan and Gilmer, Justin and Steiner, Andreas and Caron, Mathilde and Geirhos, Robert and Alabdulmohsin, Ibrahim and others},
  booktitle={ICML},
  year={2023}
}

@inproceedings{LoRA,
  author       = {Edward J. Hu and
                  Yelong Shen and
                  Phillip Wallis and
                  Zeyuan Allen{-}Zhu and
                  Yuanzhi Li and
                  Shean Wang and
                  Lu Wang and
                  Weizhu Chen},
  title        = {LoRA: Low-Rank Adaptation of Large Language Models},
  booktitle    = {ICLR},
  year         = {2022},
}

@inproceedings{VPT,
  title={Visual prompt tuning},
  author={Jia, Menglin and Tang, Luming and Chen, Bor-Chun and Cardie, Claire and Belongie, Serge and Hariharan, Bharath and Lim, Ser-Nam},
  booktitle={ECCV},
  year={2022},
}

@inproceedings{TSA,
  title={Cross-domain few-shot learning with task-specific adapters},
  author={Li, Wei-Hong and Liu, Xialei and Bilen, Hakan},
  booktitle={CVPR},
  pages={7161--7170},
  year={2022}
}

@article{VTAB,
  title={A large-scale study of representation learning with the visual task adaptation benchmark},
  author={Zhai, Xiaohua and Puigcerver, Joan and Kolesnikov, Alexander and Ruyssen, Pierre and Riquelme, Carlos and Lucic, Mario and Djolonga, Josip and Pinto, Andre Susano and Neumann, Maxim and Dosovitskiy, Alexey and others},
  journal={arXiv preprint arXiv:1910.04867},
  year={2019}
}

@inproceedings{meta-dataset,
  author       = {Eleni Triantafillou and
                  Tyler Zhu and
                  Vincent Dumoulin and
                  Pascal Lamblin and
                  Utku Evci and
                  Kelvin Xu and
                  Ross Goroshin and
                  Carles Gelada and
                  Kevin Swersky and
                  Pierre{-}Antoine Manzagol and
                  Hugo Larochelle},
  title        = {Meta-Dataset: {A} Dataset of Datasets for Learning to Learn from Few
                  Examples},
  booktitle    = {ICLR},
  year         = {2020},
}

@inproceedings{Maple,
  title={Maple: Multi-modal prompt learning},
  author={Khattak, Muhammad Uzair and Rasheed, Hanoona and Maaz, Muhammad and Khan, Salman and Khan, Fahad Shahbaz},
  booktitle={CVPR},
  year={2023}
}

@inproceedings{kornblith2019better,
  title={Do better imagenet models transfer better?},
  author={Kornblith, Simon and Shlens, Jonathon and Le, Quoc V},
  booktitle={CVPR},
  year={2019}
}

@inproceedings{zhai2022scaling,
  title={Scaling vision transformers},
  author={Zhai, Xiaohua and Kolesnikov, Alexander and Houlsby, Neil and Beyer, Lucas},
  booktitle={CVPR},
  year={2022}
}

@inproceedings{islam2021broad,
  title={A broad study on the transferability of visual representations with contrastive learning},
  author={Islam, Ashraful and Chen, Chun-Fu Richard and Panda, Rameswar and Karlinsky, Leonid and Radke, Richard and Feris, Rogerio},
  booktitle={ICCV},
  pages={8845--8855},
  year={2021}
}

@inproceedings{CoCoOp,
  author       = {Kaiyang Zhou and
                  Jingkang Yang and
                  Chen Change Loy and
                  Ziwei Liu},
  title        = {Conditional Prompt Learning for Vision-Language Models},
  booktitle    = {CVPR},
  year         = {2022},
}

@article{adaptformer,
  title={Adaptformer: Adapting vision transformers for scalable visual recognition},
  author={Chen, Shoufa and Ge, Chongjian and Tong, Zhan and Wang, Jiangliu and Song, Yibing and Wang, Jue and Luo, Ping},
  journal={NeurIPS},
  year={2022}
}

@article{eurosat,
  title={Eurosat: A novel dataset and deep learning benchmark for land use and land cover classification},
  author={Helber, Patrick and Bischke, Benjamin and Dengel, Andreas and Borth, Damian},
  journal={IEEE Journal of Selected Topics in Applied Earth Observations and Remote Sensing},
  year={2019},
}

@article{sensitivity,
  title={On sensitivity of meta-learning to support data},
  author={Agarwal, Mayank and Yurochkin, Mikhail and Sun, Yuekai},
  journal={NeurIPS},
  year={2021}
}

@article{cross_validation,
  title={A survey of cross-validation procedures for model selection},
  author={Arlot, Sylvain and Celisse, Alain},
  journal={arXiv preprint arXiv:0907.4728},
  year={2009}
}

@inproceedings{kohavi1995study,
  title={A study of cross-validation and bootstrap for accuracy estimation and model selection},
  author={Kohavi, Ron and others},
  booktitle={IJCAI},
  year={1995},
}

@inproceedings{stanfordcars,
  title={3d object representations for fine-grained categorization},
  author={Krause, Jonathan and Stark, Michael and Deng, Jia and Fei-Fei, Li},
  booktitle={ICCV workshops},
  year={2013}
}

@misc{stanforderror,
  author = {Cleanlab},
  title = {Stanford Cars (cars196) Dataset contains Many Errors},
  howpublished = "\url{https://www.linkedin.com/feed/update/urn:li:activity:7067249290959589376/}",
  year = {2023}, 
}

@inproceedings{adam,
  title={Adam: A method for stochastic optimization},
  author={Kingma, Diederik P and Ba, Jimmy},
  booktitle={ICLR},
  year={2015}
}

@article{PlantD,
  title={Using deep learning for image-based plant disease detection},
  author={Mohanty, Sharada P and Hughes, David P and Salath{\'e}, Marcel},
  journal={Frontiers in plant science},
    year={2016},
}

@article{ucf101,
  title={UCF101: A dataset of 101 human actions classes from videos in the wild},
  author={Soomro, Khurram and Zamir, Amir Roshan and Shah, Mubarak},
  journal={arXiv preprint arXiv:1212.0402},
  year={2012}
}

@article{wang2025dualbranchpromptingmultimodalmachine,
      title={Dual-branch Prompting for Multimodal Machine Translation}, 
      author={Jie Wang and Zhendong Yang and Liansong Zong and Xiaobo Zhang and Dexian Wang and Ji Zhang},
  journal={arXiv e-prints},
  pages={arXiv--2507.17588},
  year={2023}
}

@article{wenzel2020hyperparameter,
  title={Hyperparameter ensembles for robustness and uncertainty quantification},
  author={Wenzel, Florian and Snoek, Jasper and Tran, Dustin and Jenatton, Rodolphe},
  journal={NeurIPS},
  year={2020}
}

@inproceedings{ImageNet_sketch,
        title={Learning Robust Global Representations by Penalizing Local Predictive Power},
        author={Wang, Haohan and Ge, Songwei and Lipton, Zachary and Xing, Eric P},
        booktitle={NeurIPS},
        year={2019}
}

@inproceedings{DTD,
  title={Describing textures in the wild},
  author={Cimpoi, Mircea and Maji, Subhransu and Kokkinos, Iasonas and Mohamed, Sammy and Vedaldi, Andrea},
  booktitle={CVPR},
  year={2014}
}

@article{CIFAR,
  title={Learning multiple layers of features from tiny images},
  author={Krizhevsky, Alex and Hinton, Geoffrey and others},
  year={2009},
  publisher={Toronto, ON, Canada}
}

@inproceedings{VGG,
  title={Automated flower classification over a large number of classes},
  author={Nilsback, Maria-Elena and Zisserman, Andrew},
  booktitle={2008 Sixth Indian conference on computer vision, graphics \& image processing},
  year={2008},
}

@misc{QuickD,
  title={The Quick, Draw! – A.I. experiment},
  author={Jonas, Jongejan and Henry, Rowley and Takashi, Kawashima and Jongmin, Kim and Nick, Fox-Gie},
  year={2016},
  howpublished={\url{quickdraw.withgoogle.com}}
}

@misc{Fungi,
  title={{FGVCx} fungi classification challenge 2018},
  author={Brigit Schroeder and Yin Cui},
  year={2018},
  howpublished={\url{github.com/visipedia/fgvcx_fungi_comp}}
}

@article{aircraft,
  title={Fine-grained visual classification of aircraft},
  author={Maji, Subhransu and Rahtu, Esa and Kannala, Juho and Blaschko, Matthew and Vedaldi, Andrea},
  journal={arXiv preprint arXiv:1306.5151},
  year={2013}
}

@inproceedings{resnet,
  title={Deep residual learning for image recognition},
  author={He, Kaiming and Zhang, Xiangyu and Ren, Shaoqing and Sun, Jian},
  booktitle={CVPR},
  year={2016}
}

@inproceedings{swin,
  title={Swin transformer: Hierarchical vision transformer using shifted windows},
  author={Liu, Ze and Lin, Yutong and Cao, Yue and Hu, Han and Wei, Yixuan and Zhang, Zheng and Lin, Stephen and Guo, Baining},
  booktitle={ICCV},
  year={2021}
}

@inproceedings{convnext,
  title={A convnet for the 2020s},
  author={Liu, Zhuang and Mao, Hanzi and Wu, Chao-Yuan and Feichtenhofer, Christoph and Darrell, Trevor and Xie, Saining},
  booktitle={CVPR},
  year={2022}
}

@inproceedings{MAE,
  title={Masked autoencoders are scalable vision learners},
  author={He, Kaiming and Chen, Xinlei and Xie, Saining and Li, Yanghao and Doll{\'a}r, Piotr and Girshick, Ross},
  booktitle={CVPR},
  year={2022}
}

@inproceedings{IBOT,
  author       = {Jinghao Zhou and
                  Chen Wei and
                  Huiyu Wang and
                  Wei Shen and
                  Cihang Xie and
                  Alan L. Yuille and
                  Tao Kong},
  title        = {Image {BERT} Pre-training with Online Tokenizer},
  booktitle    = {ICLR},
  year         = {2022},
}

@inproceedings{EsViT,
  author       = {Chunyuan Li and
                  Jianwei Yang and
                  Pengchuan Zhang and
                  Mei Gao and
                  Bin Xiao and
                  Xiyang Dai and
                  Lu Yuan and
                  Jianfeng Gao},
  title        = {Efficient Self-supervised Vision Transformers for Representation Learning},
  booktitle    = {ICLR},
  year         = {2022},
}

@inproceedings{linear_probe,
  title={Colorful image colorization},
  author={Zhang, Richard and Isola, Phillip and Efros, Alexei A},
  booktitle={ECCV},
  year={2016},
}

@inproceedings{BitFit,
  author       = {Elad Ben Zaken and
                  Yoav Goldberg and
                  Shauli Ravfogel},
  title        = {BitFit: Simple Parameter-efficient Fine-tuning for Transformer-based
                  Masked Language-models},
  booktitle    = {ACL},
  year         = {2022},
}

@article{SSF,
  title={Scaling \& shifting your features: A new baseline for efficient model tuning},
  author={Lian, Dongze and Zhou, Daquan and Feng, Jiashi and Wang, Xinchao},
  journal={NeurIPS},
  year={2022}
}

@inproceedings{Adapter,
  title={Parameter-efficient transfer learning for NLP},
  author={Houlsby, Neil and Giurgiu, Andrei and Jastrzebski, Stanislaw and Morrone, Bruna and De Laroussilhe, Quentin and Gesmundo, Andrea and Attariyan, Mona and Gelly, Sylvain},
  booktitle={ICML},
  year={2019},
}

@article{zhang2023channel,
  title={From Channel Bias to Feature Redundancy: Uncovering the" Less is More" Principle in Few-Shot Learning},
  author={Zhang, Ji and Luo, Xu and Gao, Lianli and Zou, Difan and Shen, Hengtao and Song, Jingkuan},
  journal={arXiv e-prints},
  pages={arXiv--2310},
  year={2023}
}

@article{zhang2025closer,
  title={A Closer Look at Conditional Prompt Tuning for Vision-Language Models},
  author={Zhang, Ji and Wu, Shihan and Gao, Lianli and Song, Jingkuan and Sebe, Nicu and Shen, Heng Tao},
  journal={arXiv preprint arXiv:2506.23856},
  year={2025}
}

@article{zhang2025reliable,
  title={Reliable Few-shot Learning under Dual Noises},
  author={Zhang, Ji and Song, Jingkuan and Gao, Lianli and Sebe, Nicu and Shen, Heng Tao},
  journal={IEEE Transactions on Pattern Analysis and Machine Intelligence},
  year={2025},
  publisher={IEEE}
}

@inproceedings{kgcoop,
  title={Visual-language prompt tuning with knowledge-guided context optimization},
  author={Yao, Hantao and Zhang, Rui and Xu, Changsheng},
  booktitle={CVPR},
  year={2023}
}

@inproceedings{ProGrad,
  title={Prompt-aligned gradient for prompt tuning},
  author={Zhu, Beier and Niu, Yulei and Han, Yucheng and Wu, Yue and Zhang, Hanwang},
  booktitle={ICCV},
  year={2023}
}

@article{zhang2022progressive,
  title={Progressive meta-learning with curriculum},
  author={Zhang, Ji and Song, Jingkuan and Gao, Lianli and Liu, Ye and Shen, Heng Tao},
  journal={IEEE Transactions on Circuits and Systems for Video Technology},
  volume={32},
  number={9},
  pages={5916--5930},
  year={2022},
  publisher={IEEE}
}

@article{chavan2023one,
  title={One-for-All: Generalized LoRA for Parameter-Efficient Fine-tuning},
  author={Chavan, Arnav and Liu, Zhuang and Gupta, Deepak and Xing, Eric and Shen, Zhiqiang},
  journal={arXiv preprint arXiv:2306.07967},
  year={2023}
}

@inproceedings{ImageNet,
  title={Imagenet: A large-scale hierarchical image database},
  author={Deng, Jia and Dong, Wei and Socher, Richard and Li, Li-Jia and Li, Kai and Fei-Fei, Li},
  booktitle={CVPR},
  year={2009},
}

@inproceedings{domainnet,
  title={Moment matching for multi-source domain adaptation},
  author={Peng, Xingchao and Bai, Qinxun and Xia, Xide and Huang, Zijun and Saenko, Kate and Wang, Bo},
  booktitle={ICCV},
  year={2019}
}

@inproceedings{luo2023closer,
  title={A closer look at few-shot classification again},
  author={Luo, Xu and Wu, Hao and Zhang, Ji and Gao, Lianli and Xu, Jing and Song, Jingkuan},
  booktitle={International Conference on Machine Learning},
  pages={23103--23123},
  year={2023},
  organization={PMLR}
}

@article{zhang2022few,
  title={Few-shot learning for fine-grained emotion recognition using physiological signals},
  author={Zhang, Tianyi and El Ali, Abdallah and Hanjalic, Alan and Cesar, Pablo},
  journal={IEEE Transactions on Multimedia},
  volume={25},
  pages={3773--3787},
  year={2022},
  publisher={IEEE}
}

@article{tian2022adversarial,
  title={An adversarial meta-training framework for cross-domain few-shot learning},
  author={Tian, Pinzhuo and Xie, Shaorong},
  journal={IEEE Transactions on Multimedia},
  volume={25},
  pages={6881--6891},
  year={2022},
  publisher={IEEE}
}

@article{wu2024fine,
  title={Fine-tuning for few-shot image classification by multimodal prototype regularization},
  author={Wu, Qianhao and Qi, Jiaxin and Zhang, Dong and Zhang, Hanwang and Tang, Jinhui},
  journal={IEEE Transactions on Multimedia},
  volume={26},
  pages={8543--8556},
  year={2024},
  publisher={IEEE}
}

@inproceedings{caltech,
  title={Learning generative visual models from few training examples: An incremental bayesian approach tested on 101 object categories},
  author={Fei-Fei, Li and Fergus, Rob and Perona, Pietro},
  booktitle={2004 conference on computer vision and pattern recognition workshop},
  pages={178--178},
  year={2004},
  organization={IEEE}
}

@inproceedings{oxfordcat,
  title={Cats and dogs},
  author={Parkhi, Omkar M and Vedaldi, Andrea and Zisserman, Andrew and Jawahar, CV},
  booktitle={2012 IEEE conference on computer vision and pattern recognition},
  pages={3498--3505},
  year={2012},
  organization={IEEE}
}

@inproceedings{bossard2014food,
  title={Food-101--mining discriminative components with random forests},
  author={Bossard, Lukas and Guillaumin, Matthieu and Van Gool, Luc},
  booktitle={European conference on computer vision},
  pages={446--461},
  year={2014},
  organization={Springer}
}

@article{zhang2021deep,
  title={Deep-IRTarget: An automatic target detector in infrared imagery using dual-domain feature extraction and allocation},
  author={Zhang, Ruiheng and Xu, Lixin and Yu, Zhengyu and Shi, Ye and Mu, Chengpo and Xu, Min},
  journal={IEEE Transactions on Multimedia},
  volume={24},
  pages={1735--1749},
  year={2021},
  publisher={IEEE}
}

@article{zhang2023differential,
  title={Differential feature awareness network within antagonistic learning for infrared-visible object detection},
  author={Zhang, Ruiheng and Li, Lu and Zhang, Qi and Zhang, Jin and Xu, Lixin and Zhang, Baomin and Wang, Binglu},
  journal={IEEE Transactions on Circuits and Systems for Video Technology},
  volume={34},
  number={8},
  pages={6735--6748},
  year={2023},
  publisher={IEEE}
}

@article{zhang2024part,
  title={Part-aware correlation networks for few-shot learning},
  author={Zhang, Ruiheng and Tan, Jinyu and Cao, Zhe and Xu, Lixin and Liu, Yumeng and Si, Lingyu and Sun, Fuchun},
  journal={IEEE Transactions on Multimedia},
  volume={26},
  pages={9527--9538},
  year={2024},
  publisher={IEEE}
}

@article{zhang2025benchmark,
  title={A benchmark and frequency compression method for infrared few-shot object detection},
  author={Zhang, Ruiheng and Yang, Biwen and Xu, Lixin and Huang, Yan and Xu, Xiaofeng and Zhang, Qi and Jiang, Zhizhuo and Liu, Yu},
  journal={IEEE Transactions on Geoscience and Remote Sensing},
  year={2025},
  publisher={IEEE}
}

\begin{IEEEbiography}[{\includegraphics[width=1in,height=1.25in,clip]{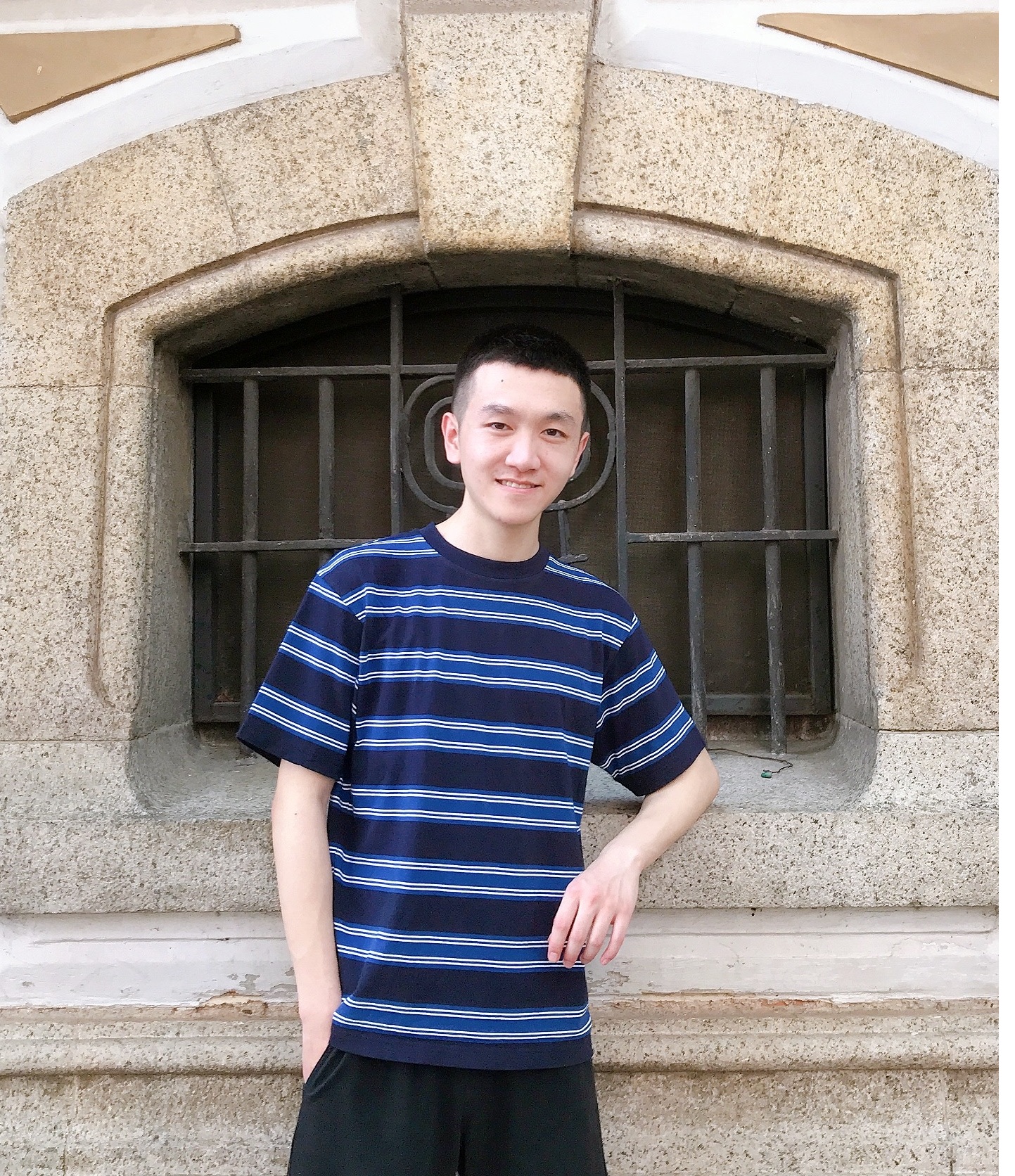}}]{Xu Luo} is a Ph.D. candidate at University of Electronic Science and Technology of China, advised by Prof. Jingkuan Song. His research interests include Computer Vision, LLMs, and Robotics. 
He has published multiple papers on top-tier machine learning conferences including NeurIPS, ICML, ICLR, CVPR, ICCV, CoRL, etc. 
% Google scholar homepage: \url{https://scholar.google.com/citations?hl=en&user=F9dKrEwAAAAJ&view_op=list_works&sortby=pubdate}.
\end{IEEEbiography}

\begin{IEEEbiography}[{\includegraphics[width=1in,height=1.25in,clip]{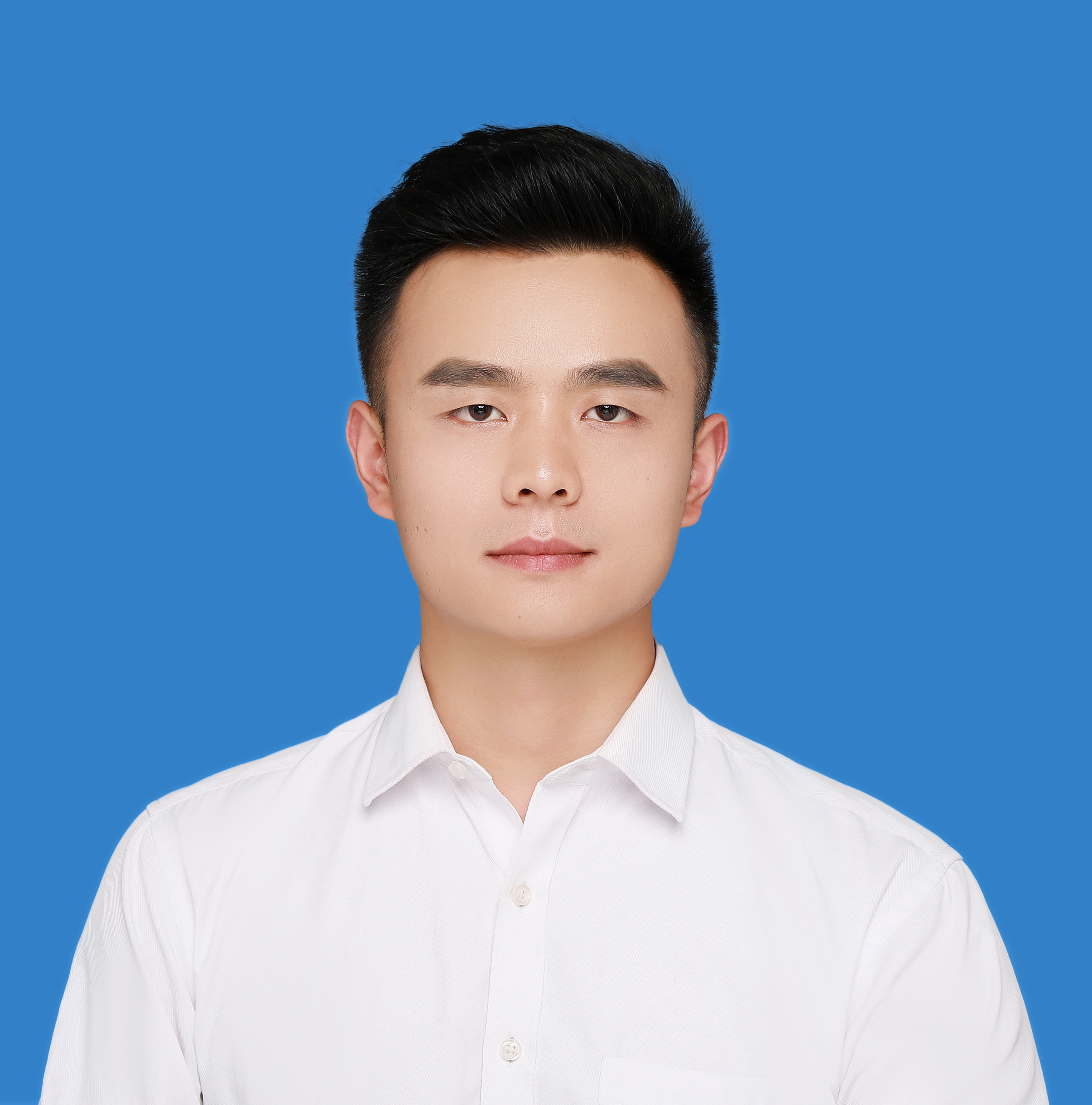}}]{Ji Zhang} is an Assistant Professor with the School of Computing and Artificial Intelligence, Southwest Jiaotong University, China. He obtained his PhD degree from University of Electronic Science and Technology of China in 2024, under the supervision of Prof. Jingkuan Song. His research interests include few-shot learning, transfer learning and robotics. 
He has published over ten papers on top conferences/journals, such as CVPR, ICCV, ICML, ICLR, TPAMI, IJCV and TIP. 
% He is/was a reviewer of CVPR'25, CVPR'24, ECCV'24, ACM MM'22-24, AAAI'23, IEEE TIP, IEEE TMM, IEEE TCSVT, and etc. 
% Google scholar homepage: \url{https://scholar.google.com/citations?hl=en&user=F9dKrEwAAAAJ&view_op=list_works&sortby=pubdate}.
\end{IEEEbiography}

\begin{IEEEbiography}[{\includegraphics[width=1in,height=1.25in,clip]{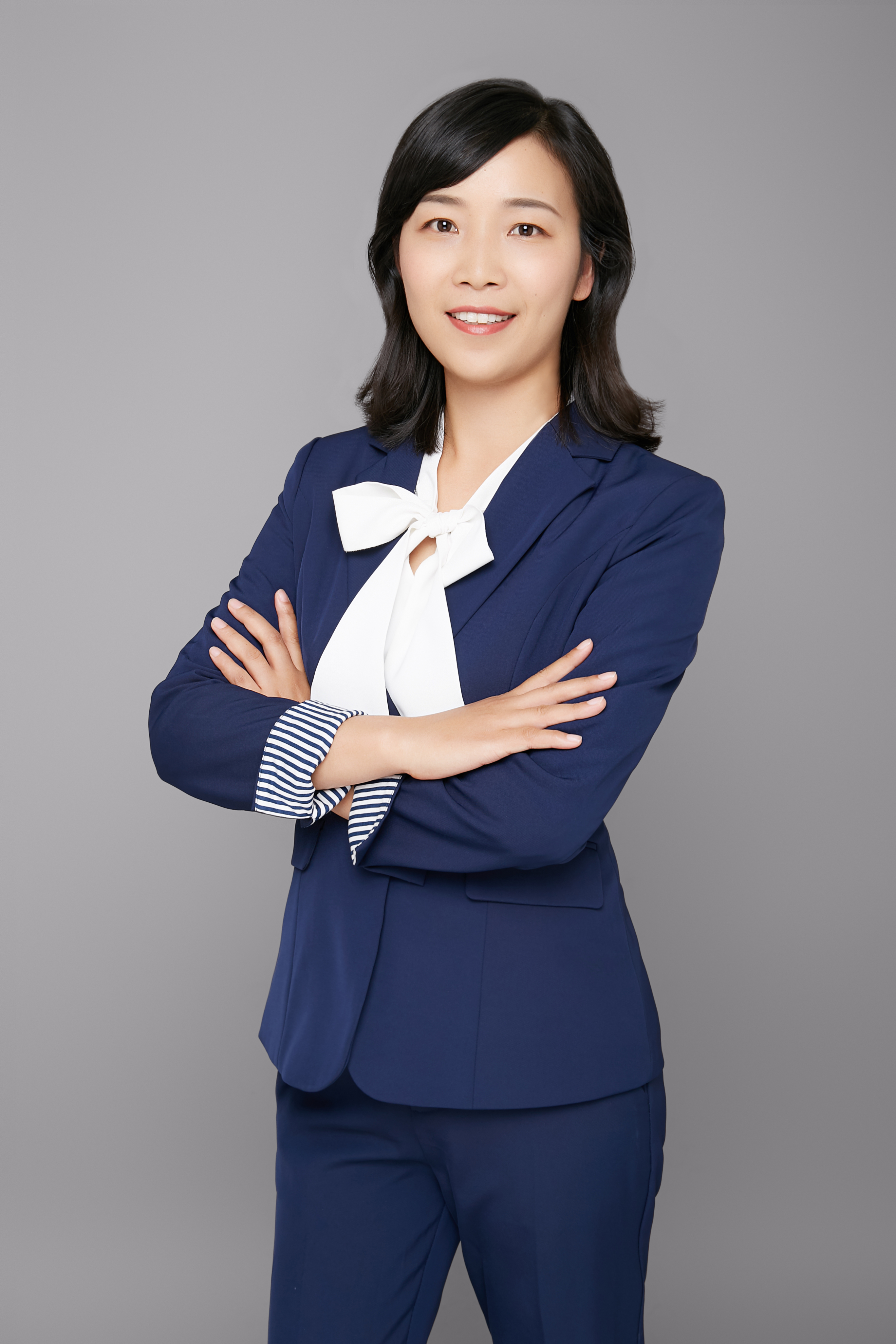}}]{Lianli Gao} is a professor with the School of Computer Science and Engineering, University of Electronic Science and Technology of China. She obtained her PhD degree in Information Technology from The University of Queensland (UQ), Australia, under the supervision of Prof. Jane Hunter and Prof. Michael Bruenig. Her research ranges from Semantic Web, Machine Learning, Deep Learning, Computer Vision (Images and Videos), NLP, Knowledge Reasoning, Knowledge and the related practical applications etc. Specifically, she is mainly focusing on integrating Natural Language for Visual Content Understanding. She has  the winner of the IEEE Transactions on Multimedia 2020 Prize Paper Award, Best Student Paper Award in Australian Database Conference (2017, Australia), IEEE TCMC Rising Star Award 2020 and ALIBABA Academic Young Fellow. She is an Associate Editor of IEEE TMM.
% Google scholar homepage: \url{https://scholar.google.com/citations?user=zsm2dpYAAAAJ&hl=en}.
\end{IEEEbiography}

\begin{IEEEbiography}[{\includegraphics[width=1in,height=1.25in,clip]{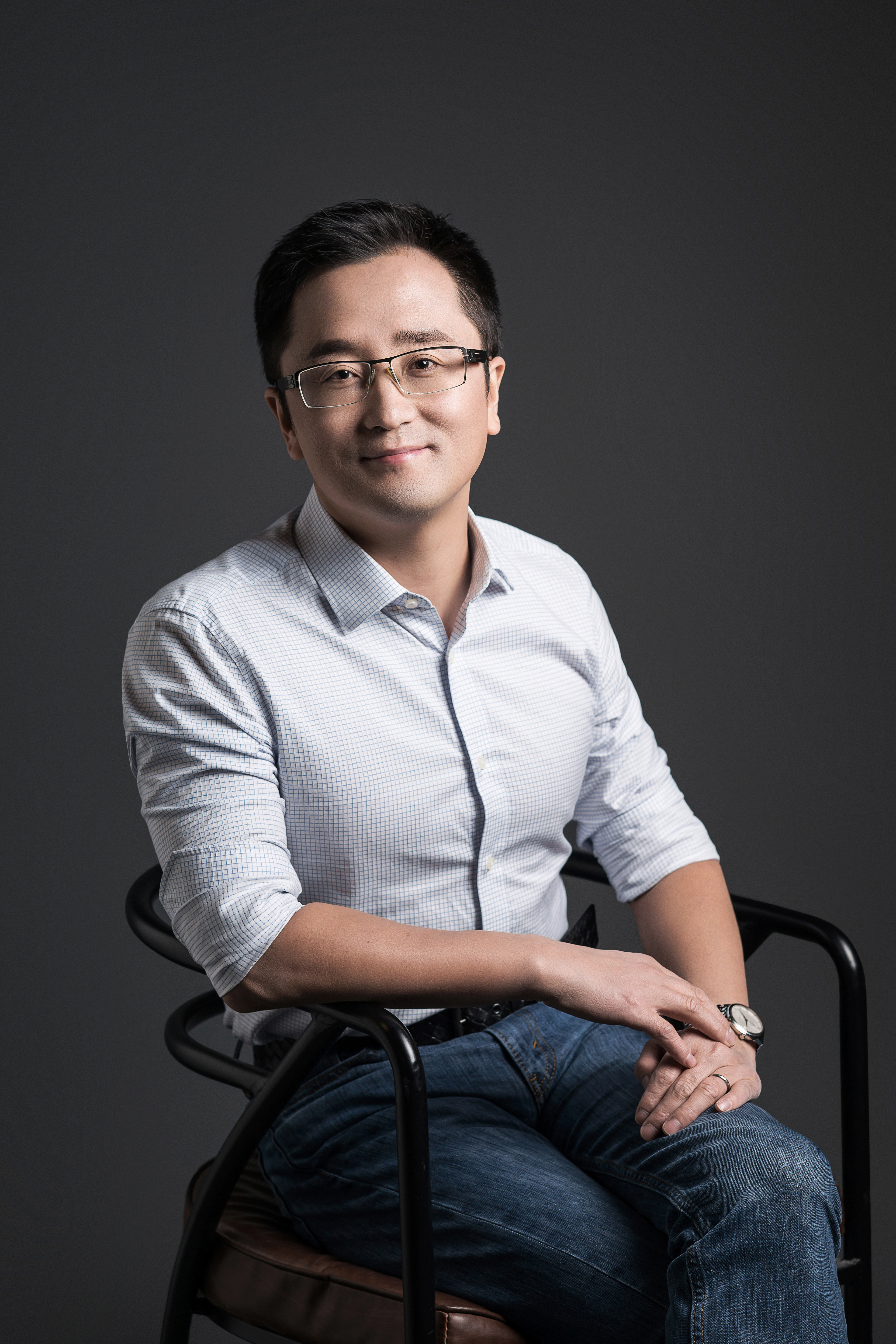}}]{Heng Tao Shen} is a professor with the School of Computer Science and Technology, Tongji University, China. 
He obtained his BSc with 1st class Honours and PhD from Department of Computer Science, National University  of  Singapore in 2000 and 2004 respectively.
His current research interests include multimedia search, computer vision, artificial intelligence, and big data management. He has published 300+ peer-reviewed papers and received 7 best paper awards from international conferences, including the Best Paper Award from ACM Multimedia 2017 and Best Paper Award-Honourable Mention from ACM SIGIR 2017. He has served as General Co-chair for ACM Multimedia 2021 and TPC Co-Chair for ACM Multimedia 2015, and is an Associate Editor of ACM Trans. of Data Science (TDS), IEEE Trans. on Image Processing (TIP), IEEE Trans. on Multimedia (TMM), and IEEE Trans. on Knowledge and Data Engineering (TKDE). He is a Fellow of ACM/IEEE/OSA.
% Google scholar homepage: \url{https://scholar.google.com/citations?user=krryaDkAAAAJ&hl=en}.
\end{IEEEbiography}

\begin{IEEEbiography}[{\includegraphics[width=1in,height=1.25in,clip]{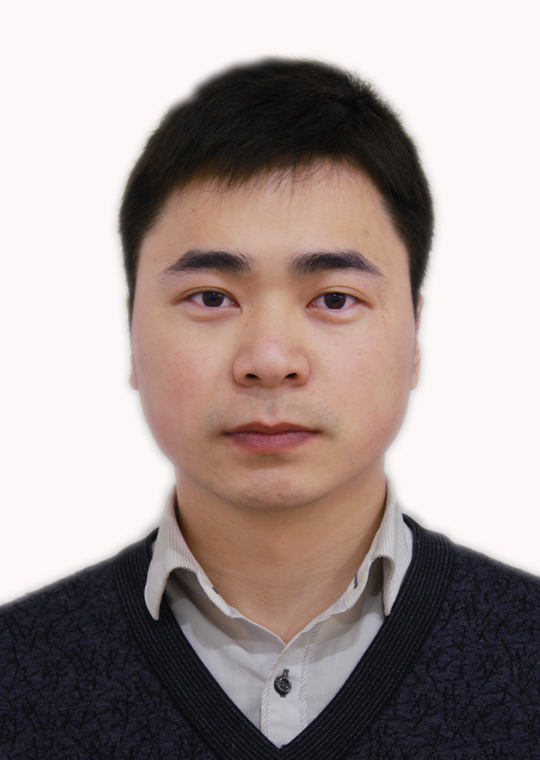}}]{Jingkuan Song} is a professor with the School of Computer
Science and Technology, Tongji University, China. 
He joined Columbia University as a Postdoc Research Scientist (2016-2017), and University of Trento as a Research Fellow (2014-2016). He obtained his PhD degree in 2014 from The University of Queensland (UQ), Australia. His research interest includes large-scale multimedia retrieval, LLMs and deep learning techniques. He was the winner of the Best Paper Award in ICPR (2016, Mexico), Best Student Paper Award in Australian Database Conference (2017, Australia), and Best Paper Honorable Mention Award (2017, Japan). He is an AE of IEEE TMM and ACM TOMM.
% and he is/was AC/SPC/PC member of CVPR'18-21, MM'18-21, AAAI'18-21, etc. 
% Google scholar homepage: \url{https://scholar.google.com/citations?user=F5Zy9V4AAAAJ&hl=en}.
\end{IEEEbiography}

\vfill

\end{document}